%% file: main.tex
\documentclass[10pt,twocolumn,letterpaper]{article}

\usepackage{iccv}              % To produce the CAMERA-READY version
\usepackage{multirow}
\usepackage{amsmath}
\usepackage{balance}
\usepackage{makecell}
\usepackage{stfloats}
\newtheorem{claim}{Claim}
\newtheorem{proof}{Proof}
\usepackage{graphicx}
\usepackage{amssymb}
\usepackage{booktabs}
\usepackage{color}
\usepackage{colortbl}


\definecolor{iccvblue}{rgb}{0.21,0.49,0.74}
\usepackage[pagebackref,breaklinks,colorlinks,allcolors=iccvblue]{hyperref}

\def\paperID{8021} % *** Enter the Paper ID here
\def\confName{ICCV}
\def\confYear{2025}

\title{VaLiD: Mitigating the Hallucination of Large Vision-Language Models by Visual Layer Fusion Contrastive Decoding}

\author{Jiaqi Wang\\
Beijing Jiaotong University\\
Beijing, China\\
{\tt\small jiaqiwang@bjtu.eud.cn}
\and
Yifei Gao\\
Beijing Jiaotong University\\
Beijing, China\\
{\tt\small yifeigao@bjtu.eud.cn}
\and
Jitao Sang\\
Beijing Jiaotong University\\
Beijing, China\\
{\tt\small jtsang@bjtu.eud.cn}
}

\begin{document}
\maketitle
\input{0_abstract}    
\input{1_intro}
\input{2_formatting}

\input{3_finalcopy}
{
    \small
    \bibliographystyle{ieeenat_fullname}
    \bibliography{main}
}

\input{X_suppl}

\end{document}

%% file: 0_abstract.tex
\begin{abstract}
Large Vision-Language Models (LVLMs) have demonstrated remarkable capabilities in multimodal task reasoning. However, they often generate responses that appear plausible yet do not accurately reflect the visual content, a phenomenon known as hallucination. Recent approaches have introduced training-free methods to mitigate hallucinations by adjusting the decoding strategy during the inference stage, typically attributing hallucinations to the language model itself. Our analysis, however, reveals that distortions in the visual encoding process significantly affect the model's reasoning capabilities. Specifically, earlier visual layers may retain key features but gradually distort as the information propagates toward the output layer. Building on these insights, we propose a novel hallucination-mitigation method from the visual encoding perspective: \textbf{V}isu\textbf{a}l \textbf{L}ayer Fus\textbf{i}on Contrastive \textbf{D}ecoding (\textbf{VaLiD}). This method utilizes uncertainty to guide the visual layer selection, correcting distortions in the visual encoding process and thereby enhancing the reliability of the generated content. Experimental results demonstrate the effectiveness of VaLiD in mitigating hallucinations across various benchmarks, achieving state-of-the-art performance when compared to baseline methods. Codes are available at \href{https://github.com/RicardoLuL/VaLiD_LVLMs_hallucinations}{Github}.
\end{abstract}

%% file: 1_intro.tex
\section{Introduction}\label{sec:intro}

\noindent Large Vision-Language Models (LVLMs) have demonstrated remarkable capabilities~\cite{instructblip,visualinstructiontuning, qwenvl} in multimodal understanding and reasoning tasks, including visual question answering~\cite{vqa, aokvqa}, cross-modal retrieval \cite{CrossmodalRetrieval}, and visual reasoning \cite{gqa}, marking them as critical milestones on the path toward Artificial General Intelligence (AGI) \cite{AGI}. Despite their impressive performance on a wide range of tasks, LVLMs still struggle with the persistent issue of hallucination~\cite{introhal1,introhal2}. This problem substantially weakens the reliability of LVLMs, presenting significant obstacles to their deployment in real-world applications~\cite{introhal3,introhal4}.

\input{Introduction}

Researchers have recently proposed various methods to address the hallucination issue in LVLMs. Most of these approaches aim to mitigate hallucinations by applying additional post-training steps~\cite{LRV-Instruction,fine-tuning1,fine-tuning2,fine-tuning3}. Other methods employ post-hoc correction during the inference phase, often relying on an auxiliary revision model~\cite{post3,post4} or manual-crafted correction pipeline~\cite{post1,post2}. However, these approaches typically face challenges such as high computational costs, complex implementation, and limited scalability. Recently, a novel paradigm focused on modifying the decoding strategy during the inference phase~\cite{decodingVCD,decodingICD,decodingritual,MARINE,decodingclip}, significantly reducing implementation costs. These methods generally assume that the language model is the primary source of hallucinations, suggesting that generated hallucinated content is mainly influenced by language priors or statistical biases~\cite{survey3, decodingVCD, decodingICD, IBD, languagepri5, languagepri6}, leading to discrepancies between the generated output and the visual content. However, tracing back to the origin of hallucinations, it is crucial to recognize that accurate and comprehensive visual input is a necessary condition to ensure correct model reasoning. Building on this insight, we propose a bold conjecture: \textit{Could hallucinations in LVLMs stem from underlying issues within the visual encoding process?}

Taking the visual counting task as an example, where the hallucination of LVLM is closely tied to its visual perception capabilities. As shown in Figure \ref{fig:intro}, LVLM generates incorrect answers when relying on the features from the standard visual output layer. In contrast, utilizing features from earlier visual hidden layers leads to correct inferences. This observation indicates that although the vision encoder captures key features in earlier layers, this information may become distorted as it propagates through the network, ultimately resulting in errors in the final decoding process.

% We identify this phenomenon as \textbf{Visual Encoding Distortion}, and observe it across different LVLMs (see Table \ref{tab:sec3.1-edr}). This finding suggests that hallucinations in LVLMs may originate at the initial stage of visual perception. To address this challenge, we propose Visual-Layer Fusion Contrastive Decoding (VaLiD), a novel hallucination mitigation method from vision encoder perspective. Specifically, we try to identify and leverage information from the layers where visual distortion occurs. In order to correct the adverse effects of inaccurate visual information, we employ a contrastive decoding approach, which compares a reference distribution, constructed from the selected distorted layers, with the original distribution generated by the standard visual output layer. This mechanism enhances the probability of correct decoding, thereby improving the reliability of the generated content.

We identify this phenomenon as \textbf{Visual Encoding Distortion}, and observe it across different LVLMs (see Table \ref{tab:sec3.1-edr}). This finding suggests that hallucinations in LVLMs may originate at the initial stage of visual perception. To address this challenge, we propose Visual-Layer Fusion Contrastive Decoding (VaLiD), a novel hallucination mitigation method from vision encoder perspective. Specifically, we try to identify and utilize information from the distorted visual layers. We then adopt a contrastive decoding approach to mitigate the adverse effects of inaccurate visual information, thereby enhancing decoding accuracy and improving the reliability of the generated content.

Our contributions are summarized as follows: (1) We emphasize the critical role of the vision encoder in the LVLM inference stage and identify a common phenomenon, Visual Encoding Distortion, observed across different LVLMs. (2) We analyze the impact of visual encoding distortion on model hallucinations, which motivates us to introduce VaLiD, the first visual-centric hallucination mitigation approach. (3) Extensive experiments demonstrate that VaLiD effectively reduces hallucinations and enhances the reliability of generated content, achieving state-of-the-art performance compared to baseline methods.

\section{Related Works}\label{related_work}
\noindent \textbf{Hallucination Mitigation in LVLMs}. In Large Vision-Language Models (LVLMs), content generated that is inconsistent with visual input is referred to as model hallucination. Various methods have been proposed to address this issue. Some approaches involve additional training steps to enhance the language model’s ability to interpret visual tokens, utilizing techniques such as auxiliary supervision~\cite{2.1_11HALC, LRV-Instruction} and reinforcement learning~\cite{2.1_4}. However, these methods inevitably rely on a large amount of annotated data and high training costs. Meanwhile, some studies aim to mitigate hallucinations during the inference stage. Certain approaches~\cite{post2,2.1_10} apply post-hoc corrections to address hallucinated content; however, their effectiveness is often constrained by reliance on auxiliary models or manual-crafted correction pipeline. Other methods~\cite{decodingVCD,decodingICD,decodingritual,decodingm3id,decodingclip,2.1_11HALC,OPERA} intervene in the decoding strategy during token generation in the inference phase, avoiding additional training or auxiliary modules, thereby reducing computational costs. However, these approaches generally attribute hallucinations to statistical biases or prior knowledge within the language model, with limited attention to the impact of the visual encoding process on hallucinations. Our method addresses this gap by focusing on the visual encoding process, tracing hallucinations to their origins in the vision encoder, and mitigating them through corrections in the visual output layer.

\vspace{0.15cm}
\noindent \textbf{Contrastive Decoding}: Contrastive decoding was initially developed to optimize the output probability distribution of language models \cite{2.2_1,2.2_2,2.2_3} and has recently been applied in LVLMs, where it improves model outputs by contrasting the conditional probabilities of responses to original versus reference inputs. Several methods have been employed to construct effective reference distributions, such as introducing uncertainty in inputs (\textit{e.g.}, VCD \cite{decodingVCD}, ICD \cite{decodingICD}, \textit{etc.}), contrastive models (\textit{e.g.}, IBD \cite{IBD}, M3ID \cite{decodingm3id}, \textit{etc.}), and data augmentation (\textit{e.g.}, RITUAL \cite{decodingritual}, VACoDe\cite{kim2024vacode}, \textit{etc.}). Additionally, work represented by DoLa \cite{DOLA} explores leveraging knowledge across different hidden layers of language models, effectively enhancing the model's reasoning capabilities on factual knowledge by contrasting information across layers. 

%Our method, while also utilizing hidden layer information, specifically focuses on dealing with the potential distortions of visual information when it is transmitted across layers within the vision encoder. By applying contrastive techniques, our approach mitigates the negative effects of these distortions on visual information, thereby effectively reducing hallucinations in LVLMs. 

%% file: Introduction.tex
\begin{figure}[t!]
    \centering
    \vspace{-1mm}
    \includegraphics[width=1\linewidth]{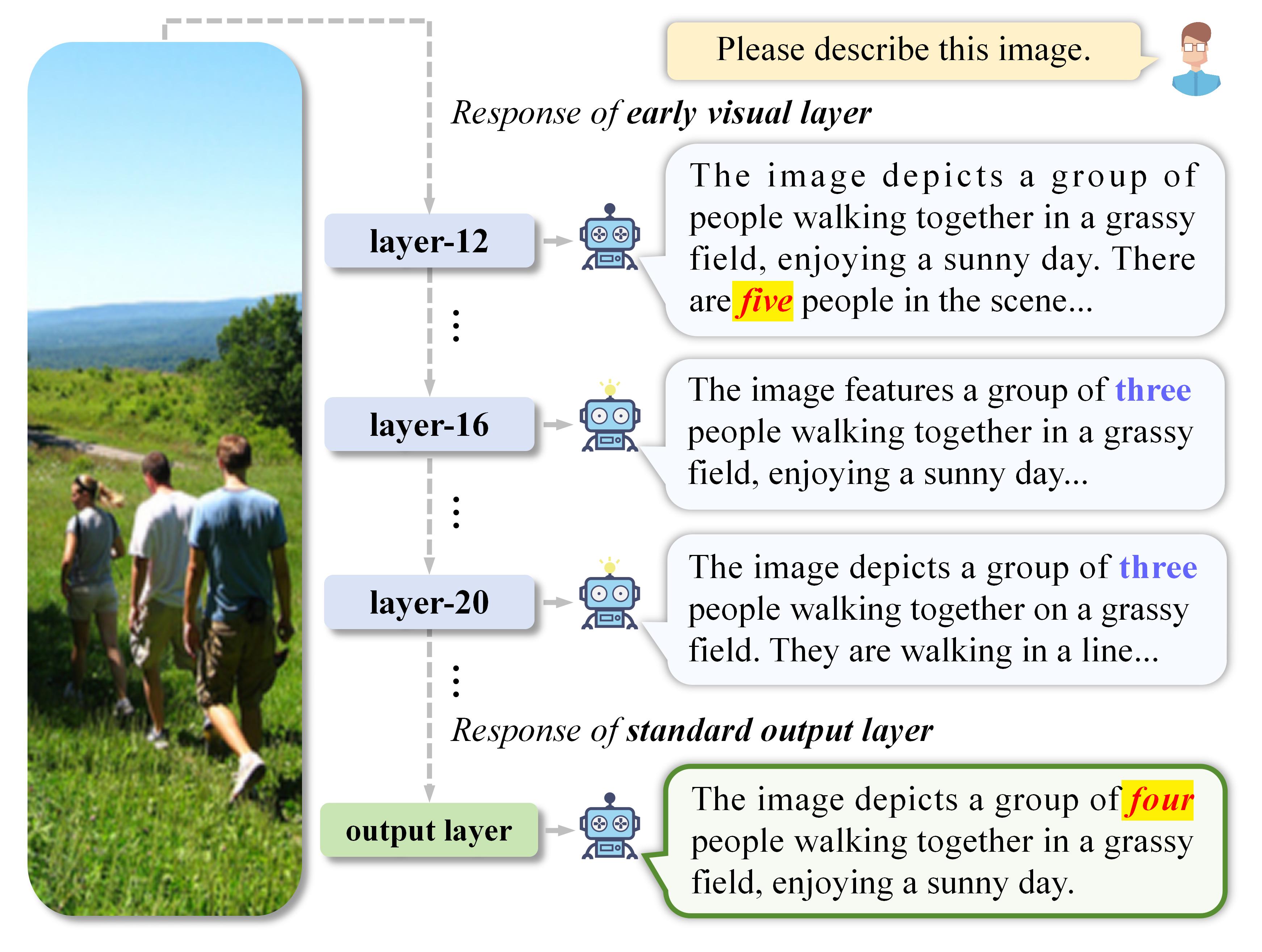}
    \caption{An example demonstrates where the LVLM provides a hallucinated response based on the features from standard visual output layer, yet correctly identifies the number of people in the image when relying on features from other visual hidden layers. The correct answer and hallucination are highlighted in \textcolor{blue!60}{\textbf{blue}} and \colorbox{yellow}{\textcolor{red}{\emph{\textbf{red}}}}, respectively.}%
    \label{fig:intro}
    \vspace{-0.25cm}
\end{figure}

%% file: 2_formatting.tex
\section{Methodology} \label{sec:Methodology}
\subsection{Visual Encoding Distortion Induces Hallucinations} \label{Sec3.1}
In general, the features of the standard visual output layer are effective for multimodal reasoning tasks. However, as shown in Figure \ref{fig:intro}, they may not be the most reliable when addressing problems prone to hallucination, for example, scenarios involving semantic conflicts such as object attributes, relationships, or existence \cite{pope}. This observation motivates further study into the impact of the visual encoding process with respect to LVLMs' hallucination behavior.

\input{table_section3_EDR}

\input{section3-finding2}

\vspace{0.15cm}
\noindent\textbf{Finding 1: Encoding distortion in visual layers causes most prediction reversals.} We examine the decoding results across all visual layers to quantify potential distortion in the visual encoding process. Specifically, we introduce the Encoding Distortion Rate (EDR) to measure the proportion of samples that are correctly decoded at early visual layers but incorrectly decoded at the standard output layer. The EDR score is defined as: $\frac{S_{i}^{+}\cap S_{so}^{-}}{S_{so}^{-}}$, where $S_{i}^{+}$ represents the set of features from the $i$-th visual layer that are correctly decoded, and $S_{so}^{-}$ denotes the set of incorrectly decoded features from the standard visual output layer. To provide a more intuitive representation of our findings, we aggregate accuracy statistics based on predefined layer groupings, referred to as ``buckets." A bucket is considered to yield correct answers if the visual features within it enable accurate decoding by the LLMs. Detailed settings are provided in the supplementary material \ref{EDR}.

As shown in Table \ref{tab:sec3.1-edr}, a significant proportion of samples can be decoded correctly at the early visual hidden layers but unexpectedly decode wrongly at the standard output layer. Especially in bucket 5 of the InstructBLIP (with layers ranging from 31 to 34), the EDR score reaches an astonishing 69.35\%. This suggests that key visual information has become distorted during the propagation towards the standard visual output layer and ultimately causing the model's prediction to shift from correct to incorrect.

\vspace{0.15cm}
\noindent\textbf{Finding 2: Uncertainty as Indicator of Encoding Distortion.} Based on the above finding, it is crucial to identify the visual layer where encoding distortion occurs. Inspired by~\cite{uncertainty2,uncertainty6}, we leverage uncertainty to analyze the transformation of visual information as it propagates through the visual hidden layers. The uncertainty of $i$-th visual layer, as defined in Eq. \ref{eq:1}, is measured by he entropy of the next token's probability distribution when LVLMs generate predictions based on its features.

\begin{equation}
\label{eq:1}
    H_{i, t} = - \sum_{y_{t} \in \mathcal{V}} P_{\theta}(y_{t} \vert v_{i}, x, y_{< t}) \;\text{log} P_{\theta}(y_{t} \vert v_{i}, x, y_{< t})
\end{equation}

\noindent where LVLM is parametrized by $P_{\theta}$, $\mathcal{V}$ is the vocabulary space, $v_{i}$ denotes the visual features of the $i$-th layer, $x$ denotes a query associated with image $v$ and $y_{< t}$ represents the generated tokens. We quantify the relationship between the layers uncertainty and corresponding decoding results in LVLMs. As shown in Figure \ref{finding2}, visual hidden layers associated with incorrect decoding tend to exhibit higher uncertainty, whereas those with correct decoding maintain relatively low uncertainty. This finding suggests that the visual encoding process is influenced by high-uncertainty layers, with these effects accumulating as information propagates through the network and becoming increasingly pronounced. Finally, this leads LVLMs to generate erroneous inferences based on distorted features.

\input{Framework}

\subsection{Visual-Layer Fusion Contrastive Decoding}
Building on previous findings, visual feature distortion originates from early layers with high uncertainty and ultimately leads to hallucinations during the LVLMs’ decoding process. To overcome this challenge, we propose \textbf{V}isu\textbf{a}l-\textbf{L}ayer Fus\textbf{i}on Contrastive \textbf{D}ecoding (\textbf{VaLiD}). As shown in Figure \ref{fig: framework}, VaLiD is a novel approach that leverages contrastive decoding to correct distortions by integrating information from multiple visual layers.

\vspace{0.15cm}
\noindent\textbf{Uncertainty-guided Visual-Layer Fusion}. Given that uncertainty can indicate information distortion within the visual encoding process, we propose leveraging early visual layers with high uncertainty to improve output correction through contrastive decoding. However, directly relying on hidden layers with the highest uncertainty could lead the model to excessively depend on potentially unstable information from a single layer \cite{DOLA,uncertainty6}. To address this, we introduce a layer fusion approach to enhance the robustness of contrastive decoding. Specifically, at each time step, we dynamically select the top-k layers with the highest uncertainty to construct a candidate set of layers. 

\begin{equation}
    \label{eq:2}
    \begin{aligned}
        \centering
        & P_{ref} = \sum_{i \in \mathcal{C}} \;\omega_{i} \cdot P_{\theta}(y_{t} \vert v_{i}, x, y_{< t}) \\
        & \omega_{i}=\dfrac{\text{exp}^{H_{i,t}}}{\sum_{k}^{\vert \mathcal{C} \vert} \text{exp}^{H_{k, t}}}
    \end{aligned}
\end{equation}

To ensure computational efficiency, we partition the visual encoder’s layers into several buckets, allowing fusion within specific layer ranges, with the optimal bucket determined by the validation set (see Sec. \ref{sec4.1} for implementation details). This dynamic layer selection strategy adaptively identifies the most suitable set of early visual layers based on the next token, enhancing the effective use of visual information across layers. Through this bucketing and selection strategy, we balance the computational efficiency and decoding performance. Finally, we represent the reference distribution of the next token as a entropy-weighted average of the distributions from earlier visual layers, as shown in Eq. \ref{eq:2}, where $\mathcal{C}$ denotes the candidate layer set.

\vspace{0.15cm}
\noindent \textbf{Contrastive Decoding}. After obtaining (1) the next token distribution conditioned on the standard output layer features, denoted as $P_{ori}$ and (2) the reference distribution, denoted as  $P_{ref}$, we compute a new probability distribution by contrasting the differences between these two distributions. The contrastive result is presented in Eq. \ref{eq:3}.

\vspace{-0.2cm}
\begin{equation}\label{eq:3}
    P_{valid} = (1+\alpha) P_{ori} - \alpha \sum_{i \in \mathcal{C}} \omega_{i} \cdot P_{\theta}(y_{t} \vert v_{i}, x, y_{< t})
\end{equation}

\noindent where $\alpha$ is a positive coefficient to control the contrast intensity. A larger $\alpha$ value indicates a stronger amplification of differences between the two distributions ($\alpha=0$ reduces to regular decoding, we set $\alpha=1$ in this paper, details please refer to the supplementary material \ref{aba-study}). However, the newly generated distribution $P_{\text{valid}}$ may mistakenly penalize valid outputs in $P_{\text{ori}}$, leading to decoding results that do not adhere to basic language standards and commonsense reasoning. To address this issue, we adopt an adaptive reliability constraint following the approach of \cite{2.1_1}. For further details, please refer to the supplementary material \ref{adaptive-reliability-constraint}. 

%% file: table_section3_EDR.tex
\begin{table}[t!]
    \centering
    \renewcommand\arraystretch{1.1}
    \setlength{\tabcolsep}{4pt}
    \resizebox{1\linewidth}{!}{%
    \begin{tabular}{c c c c c c c}
    \toprule
        \textbf{Model} & \textbf{bucket1} & \textbf{bucket2} & \textbf{bucket3} & \textbf{bucket4} & \textbf{bucket5} & \textbf{bucket6} \\ \midrule
        LLaVA-v1.5 \cite{llava1.5} & 0.00 & 0.88 & 8.85 & 18.58 & \textbf{33.63} & 30.09 \\ 
        InstructBLIP \cite{instructblip} & 16.13 & 16.13 & 33.87 & 61.29 & \textbf{69.35} & 54.84 \\ 
        Qwen-VL \cite{qwenvl} & 0.00 & 7.46 & \textbf{14.93} & 4.38 & 8.96 & 1.49 \\ \bottomrule
    \end{tabular}
    }
    \caption{Results of EDR score across three representative models using AMBER \cite{AMBER} benchmark.}
    \label{tab:sec3.1-edr}
    \vspace{-0.5cm}
\end{table}

%% file: section3-finding2.tex
\begin{figure}[t!]
    \centering
    
    \begin{minipage}{0.48\linewidth}
        \centering
        \centerline{\includegraphics[width=\textwidth]{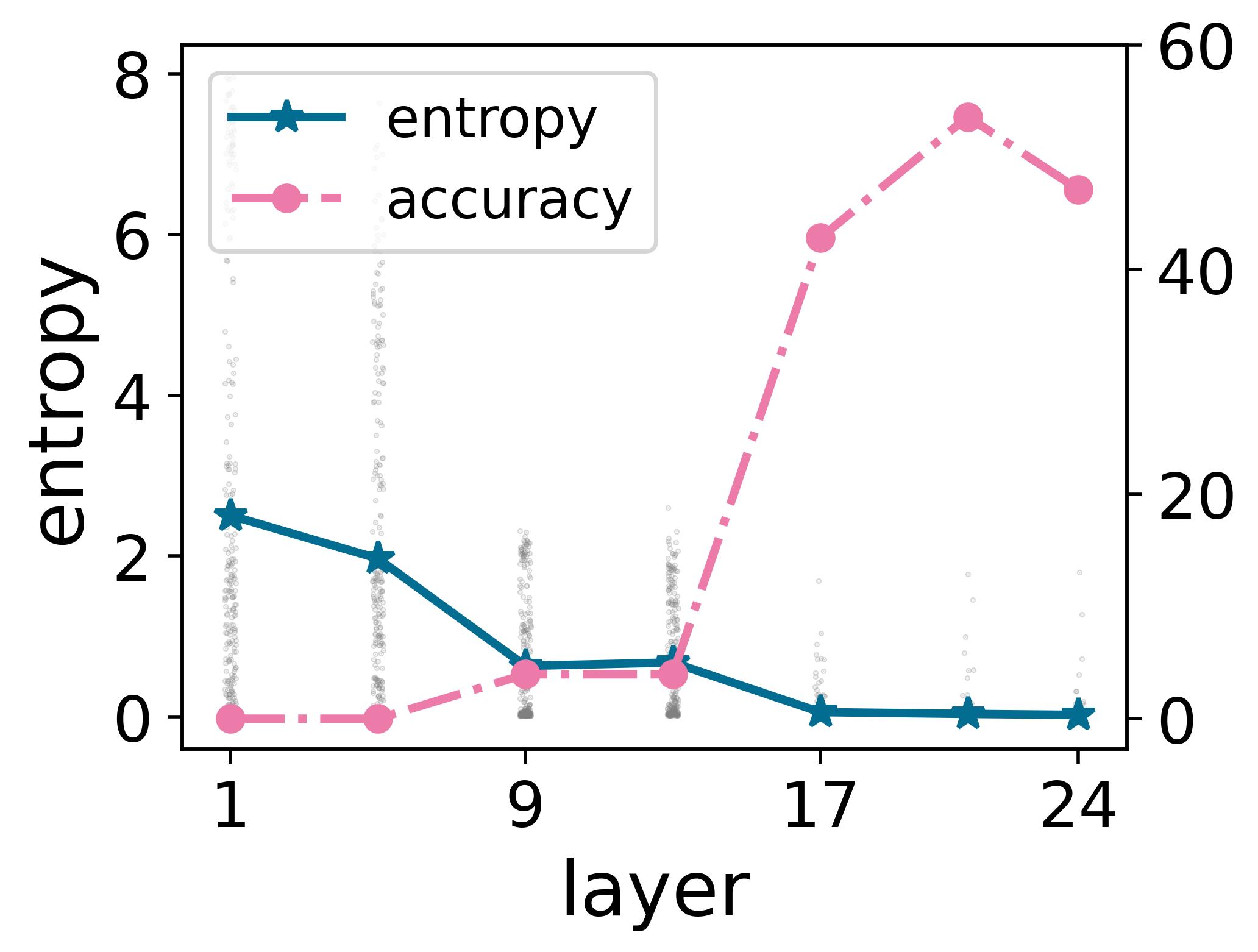}}
        \small\centerline{(a) LLaVA-v1.5}
    \end{minipage}
    %\qquad
    \begin{minipage}{0.48\linewidth}
        \centering
        \centerline{\includegraphics[width=\textwidth]{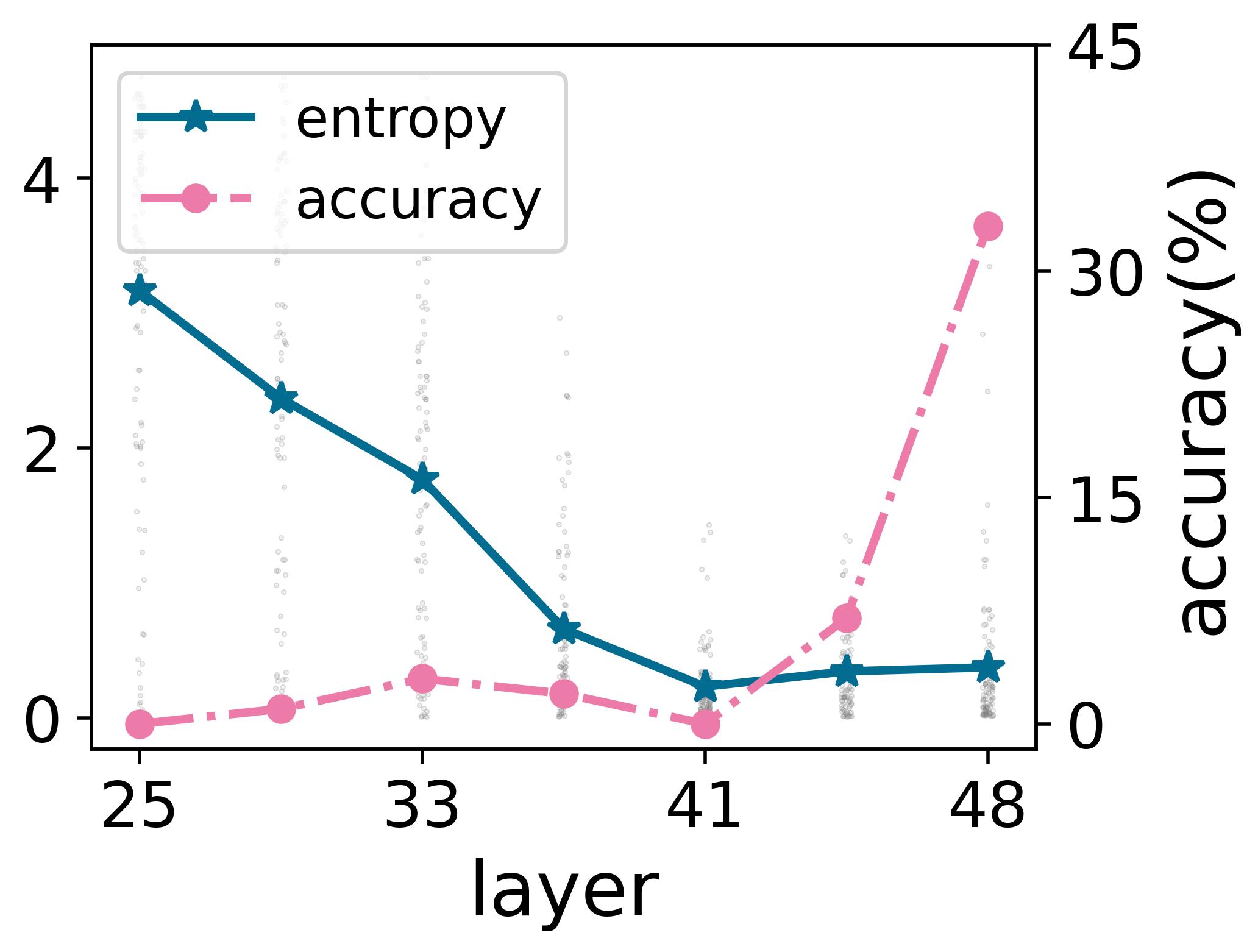}}
        \small\centerline{(b) Qwen-VL}
    \end{minipage}
    \caption{Results of layer uncertainty and decoding accuracy of all visual hidden layers. Each node on the blue curve represents the average value of entropy across the AMBER benchmark. It should be noted that LLaVA-v1.5 utilizes the feature from the penultimate layer of CLIP-ViT as the standard visual output..}
    \label{finding2}%文中引用该图片代号
    \vspace{-0.3cm}
\end{figure}

%% file: Framework.tex
\begin{figure*}[tp]
    \centering
    \includegraphics[width=1\linewidth]{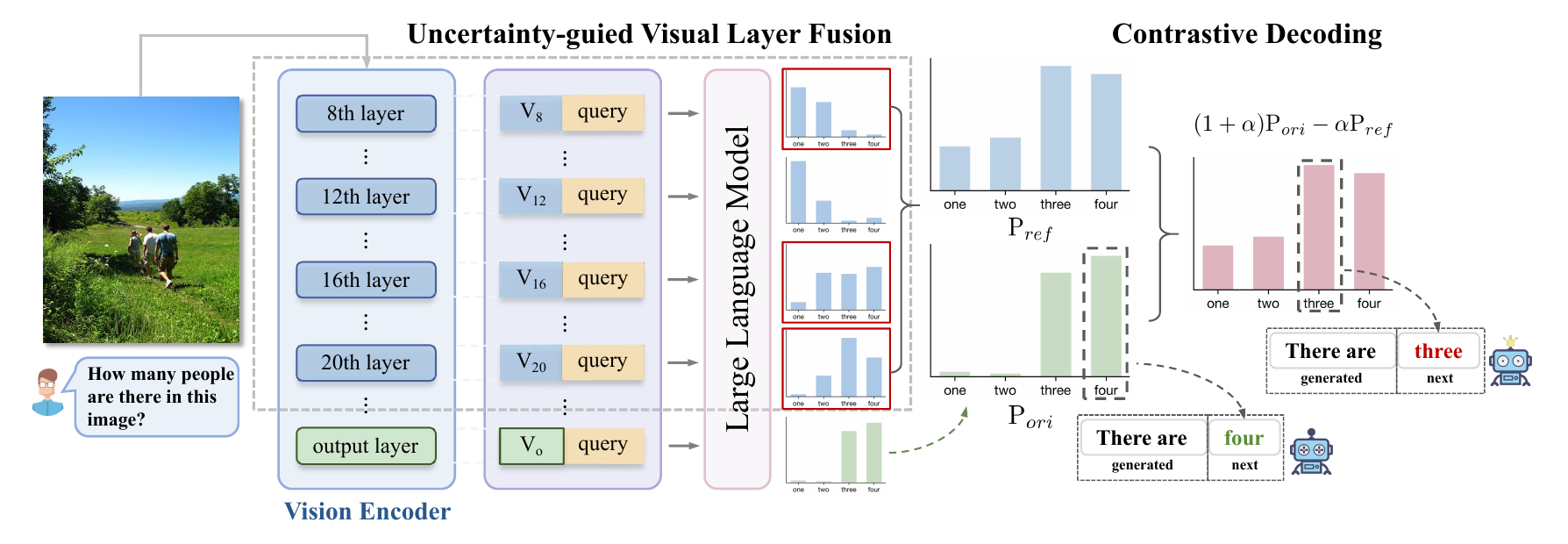}
    \vspace{-0.4cm}
    \caption{Overview of the VaLiD decoding process. At each time step, the LVLM auto-regressively samples the token $y_{t}$ based on visual input, text query, and previously generated tokens. The probability distributions represented in \textcolor{green!30}{\textbf{Green}} and \textcolor{blue!40}{\textbf{Blue}} correspond to decoding results from the standard visual output layer and early visual layers, respectively. The red boxes indicate that the probability distributions of the selected visual layers, which reflect the top-k uncertainty, will be used to calculate the reference distribution. The final corrected probability distribution, shown in \textcolor{red!40}{\textbf{Red}}, is obtained through contrastive decoding. In this case, when asked about the number of people in the image, LVLM generates the correct decoding result, ``three", instead of the incorrect ``four" from the original distribution.}
    \label{fig: framework}
    \vspace{-0.3cm}
\end{figure*}

%% file: 3_finalcopy.tex
\section{Experiment}
\subsection{Evaluation Setup}\label{sec4.1}

\input{table_pope}

\input{table_amber}

\noindent\textbf{LVLMs and Layer Selection}: In this section, we integrate VaLiD with three representative LVLMs: InstructBLIP-7B \cite{instructblip}, LLaVA-v1.5-7B \cite{llava1.5}, and Qwen-VL-7B \cite{qwenvl}. Each model utilizes Vision Transformer (ViT) as the backbone of its visual encoder but employs different pre-training strategies, such as Eva-ViT (39 layers) \cite{evavit}, CLIP (25 layers) \cite{clip}, and ViT-bigG (49 layers), where the numbers indicate the layer count of the visual encoder. To ensure computational efficiency, we do not pass all visual layers through the LLM. Specifically, we (1) partition visual layers into distinct buckets; (2) identify one bucket as the candidate layer set using the validation set; and (3) select the top k layers with the highest entropy from the best performing bucket. Detailed settings and validated results are available in the supplementary material \ref{exp-settings}.

\vspace{0.15cm}

\noindent\textbf{Datasets and Benchmarks}: We evaluate our approach on three representative hallucination benchmarks: (1) \textbf{POPE} \cite{pope}, which assesses object hallucination in LVLMs by asking them to answer ``yes/no" questions regarding object existence. This benchmark includes evaluations across three datasets: MS-COCO \cite{mscoco}, A-OKVQA \cite{aokvqa}, and GQA \cite{gqa}. Each dataset contains three subsets, random, popular, and adversarial, based on different negative sampling strategies. These subsets offer increasing levels of difficulty, providing a progressive assessment of whether LVLMs tend to generate specific objects erroneously. (2) \textbf{AMBER} \cite{AMBER}, which extends POPE by providing more fine-grained annotations, including 1,004 images and 15,200 corresponding annotations. This benchmark evaluates three distinct types of hallucinations, existence, attribute, and relational hallucinations, offering a more comprehensive assessment of LVLMs' ability to mitigate these errors. (3) \textbf{MME} \cite{mme}, a comprehensive benchmark for evaluating LVLMs, covering 14 tasks across various dimensions. Following the experimental settings in VCD \cite{decodingVCD}, we use the existence and count subsets to assess object-level hallucination, while the location and color subsets are used for the evaluation of attribute-level hallucination. Additionally, we report LVLM performance across the remaining 10 tasks to examine whether the proposed decoding strategy compromises the model's reasoning capability. Detailed metric settings are provided in the supplementary material \ref{exp-settings}.

% \vspace{0.15cm}
\noindent\textbf{Baseline Settings}: Unlike post-hoc hallucination mitigation methods, our approach directly modifies the model’s decoding strategy. Consequently, we restrict our baseline comparisons to alternative decoding strategies. Specifically, we select VCD \cite{decodingVCD}, M3ID \cite{decodingm3id}, and Ritual \cite{decodingritual} as baseline methods. Both VCD and M3ID employ a contrastive decoding strategy based on prior knowledge, assuming that hallucinations primarily stem from the language model. In contrast, Ritual maintains consistent visual representations through a data augmentation strategy. In addition, we report the evaluation results of the vanilla model, which does not incorporate any specialized decoding strategy. Detailed baseline settings are provided in supplementary material \ref{baseline-settings}. Given that language-layer decoding methods~\cite{DOLA, damo} have been explored for hallucination mitigation in LLMs, they can naturally be extended to LVLMs as well. In subsequent discussions, we compare our approach with these methods, emphasizing the necessity of a visual-centric solution for effective hallucination mitigation.

\subsection{Experimental Results}
\noindent\textbf{Results on POPE}: Table \ref{tab:pope} presents the overall performance comparison of VaLiD and other baseline methods on the POPE benchmark. Our findings lead to two key conclusions: (1) VaLiD significantly outperforms the vanilla model across all three LVLMs, demonstrating that eliminating adverse effects of visual encoding distortion effectively mitigates hallucination. (2) VaLiD outperforms those methods attributing hallucination to language prior (e.g., VCD and M3ID), suggesting that, the correctness of visual encoding is also crucial for hallucination mitigation. Moreover, when examining the impact of challenging sampling scenarios on model hallucinations, we find that VaLiD demonstrates greater stability across different sampling methods, whereas baseline methods exhibit varying degrees of performance degradation under highly correlated sampling conditions. These results suggest that VaLiD enables the model to better counteract interference from correlated data (e.g., frequently co-occurring objects) by correcting the adverse effects of inaccurate visual information, thereby allowing LVLMs to more accurately understand object relationships.

\input{mme}

\vspace{0.15cm}
\noindent\textbf{Results on AMBER}: Our evaluation on AMBER dataset demonstrates the performance of each decoding strategy in mitigating various types of hallucinations beyond object existence. As shown in Table \ref{tab:amber}, VaLiD exhibits strong competitiveness across various tasks. Notably, for questions involving object state, action and number attributes, VaLiD outperforms a series of contrastive decoding methods that rely on data augmentation, such as VCD and Ritual. This advantage arises from VaLiD’s ability to avoid additional biases introduced by data augmentation, which can alter the attribute semantics of images. For example, in VCD, adding Gaussian noise can obscure small target objects, while in Ritual, random cropping may inadvertently remove target objects. However, we observe that VaLiD performs less effectively on InstructBLIP for relation-level hallucinations, highlighting that relational reasoning remains a challenging task for InstructBLIP, with limited improvement achieved through VaLiD. Despite this, VaLiD still surpasses most baseline methods.

\vspace{0.15cm}
\noindent\textbf{Results on MME}: As shown in Figure \ref{fig:mme}, VaLiD significantly enhances performance in most perceptual and cognitive tasks across all three LVLMs, indicating its effectiveness in mitigating hallucinations without compromising the model's inherent capabilities. Notably, VaLiD even appears to boost perceptual and cognitive performance in specific tasks. For example, in the posters task, VaLiD outperforms the vanilla model by 29.59\% in LLaVA-v1.5, and in the text translation task, VaLiD surpasses the vanilla model by 40\% in Qwen-VL. However, certain cognitive tasks, such as code reasoning and numerical calculation, show better performance with the vanilla model. We believe that this may be due to the inherent challenges posed by statistical biases and language priors affecting LVLMs, which might not be fully addressed from vision encoder perspective.

\input{ablation_study}

\subsection{Ablation study}

\noindent\textbf{On layer selection}. We compare random selection with maximum entropy-based selection for visual encoder layers. The random approach uniformly picks one layer, while the entropy-based method selects the layer exhibiting the highest entropy values. The detailed settings are available in the supplementary material \ref{aba-study}. As demonstrated in Table \ref{tab:rebuttal-tabble0}, maximum entropy selection (column 2) achieves superior performance compared to random selection (column 1). This validates that high-entropy layers capture more semantically meaningful information critical for hallucination mitigation, whereas random selection risks incorporating redundant or less informative layers, resulting in less effective performance.

\vspace{0.15cm}
\noindent\textbf{On fusion strategy}. We further evaluate average-weighted versus entropy-weighted fusion strategies. The entropy-weighted method (column 4) outperforms average fusion (column 3), indicating that emphasizing high-entropy elements enhances information aggregation. By assigning weights proportional to entropy values, the fusion process preserves critical information while suppressing noise. In contrast, average weighting equally treats all elements regardless of their information content, potentially weakening key features and compromising fusion quality.

\input{table_final2}

\input{case_study}

\subsection{Discussion}
\vspace{0.15cm}
\noindent\textbf{The compatibility of VaLiD}. 
Compared to most existing methods that construct reference distributions from a language-prior perspective, VaLiD is the pioneer to build the reference distribution from a visual encoding perspective. This approach is orthogonal to previous methods, making VaLiD a versatile framework that allows other decoding strategies to be seamlessly integrated into its probability distribution output. Here, we explore the advantages of VaLiD in terms of compatibility with other methods.

As shown in Table \ref{tab:rebuttal-final2}, we compare the compatibility of VaLiD and VCD on the POPE benchmark using LLaVA-v1.5. The results indicate that when using $\mathbb{P}_{\text{valid}} \circ \mathbb{P}_{\text{vcd}}$, where VaLiD is applied after VCD, the performance surpasses that of  $\mathbb{P}_{\text{vcd}} \circ \mathbb{P}_{\text{valid}}$, where VCD is applied after VaLiD, as well as the performance of applying VCD alone and VaLiD alone. For detailed experimental settings and proofs, please refer to the supplementary material \ref{sec:appendix-compatible}. Notably, when we explore the compatibility of VCD, its performance remains nearly identical to that of standalone VCD and is even lower than that of standalone VaLiD. This finding suggests that VCD is difficult to effectively compatibly with external methods, while VaLiD shows excellent compatibility, making it a more adaptable framework for hallucination mitigation.

To further validate the universality of VaLiD's compatibility advantages, we introduce the third method Ritual to conduct compatibility comparison experiments on VaLiD and VCD. According to the results in Table \ref{tab:rebuttal-tabble2}, when the third method is incorporated, VaLiD demonstrates superior hallucination mitigation capabilities.

\input{table_ritual_vcd_valid}

\vspace{0.15cm}
\noindent\textbf{Comparison with language-layer decoding method}. 
Our study is based on a key observation: hallucinations in LVLMs do not only stem from the language priors embedded in the language model but are more likely to originate from visual encoding distortion. However, existing studies primarily attribute hallucinations to the language model, overlooking the crucial role of the visual encoding process in hallucination generation. Here, we compare our method with Dola and another Dola-like method Damo \cite{damo}. Dola \cite{DOLA} focuses on mitigating hallucinations in LLMs focuses on mitigating hallucinations in LLMs by utilizing layer-wise probability distributions. As shown in Table \ref{tab:rebuttal-tabble1}, VaLiD significantly outperforms language-level decoding methods in scenarios with high visual spurious correlations (i.e., 6\% higher in popular scenarios and 10\% higher in adversarial scenarios). This clearly demonstrates that: (i) addressing the hallucinations of LVLMs solely from the language perspective has inherent limitations, and (ii) it is crucial to consider a visual-centric solution.
% , which uses layer-wise probability distributions to adaptively select the contrastive layers.

\vspace{0.15cm}
\noindent\textbf{Case studies}: Figure \ref{fig: case_study} presents two case studies that illustrate how vanilla decoding results in object hallucinations, while VaLiD generates responses more closely aligned with the visual content. In these examples, a ``mouse" is often co-occurring with a keyboard and stacked bananas lead the model to produce hallucinated objects. In contrast, VaLiD effectively mitigates these issues. Additionally, VaLiD successfully identifies a bottle in the upper left corner in the first case, a commonly overlooked object by most LVLMs. We provide more cases in the supplementary material \ref{sec:appendix-more_case}.

\input{rebuttal1}

\section{Conclusion and Limitation}
In this paper, we introduce Visual Layer Fusion Contrastive Decoding (VaLiD), a novel, visual- centric method for mitigating hallucinations in LVLMs. Our approach highlights the critical role of the visual encoding process, showing that distorted visual information may induce hallucinations. We leverage probability distributions from hidden layers to correct distortions in the standard output layer of the visual encoder, thereby enhancing the reliability of the generated content. Extensive experimental results validate the effectiveness of VaLiD across multiple datasets and LVLMs.

\vspace{0.15cm}
\noindent\textbf{Limitation} In this work, we tried to achieve a balance between efficiency and performance through a bucketing strategy, however, the computational efficiency of VaLiD is still affected to some extent by the number of selected visual layers. Nevertheless, we believe that future work should focus on investigating the root causes of these distortions so that higher quality visual features can be extracted. In this way, a single decoding process may be sufficient to generate content that faithfully reflects the image. Additionally, future work could explore an adaptive layer contrastive decoding strategy for each token, which holds significant potential for real-world applications. These considerations further motivate us to investigate the inner workings of the vision encoder and the influence of visual information on multimodal reasoning tasks beyond hallucinations.

%% file: table_pope.tex
\begin{table*}[!ht]
    \centering
    \renewcommand\arraystretch{1.095}
    \setlength{\tabcolsep}{7pt}
    \resizebox{0.9\linewidth}{!}{%
    \begin{tabular}{c c c c c c c c c c c c c c}
    \toprule
        \multirow{2}{*}{\textbf{Dataset}} & \multirow{2}{*}{\textbf{Setup}} & \multirow{2}{*}{\textbf{Method}} & \multicolumn{3}{c}{\textbf{LLaVA-v1.5}} & ~ & \multicolumn{3}{c}{\textbf{InstructBLIP}}  & ~ &  \multicolumn{3}{c}{\textbf{Qwen-VL}}  \\ \cline{4-6}\cline{8-10}\cline{12-14}

        ~ & ~ & ~ & Acc.$\uparrow$ & F1$\uparrow$ & Yes & ~ & Acc.$\uparrow$ & F1$\uparrow$ & Yes & ~ & Acc.$\uparrow$ & F1$\uparrow$ & Yes \\ \hline
        \multirow{15}{*}{\rotatebox{90}{\textbf{\normalsize MSCOCO \cite{mscoco}}}} & \multirow{5}{*}{\textit{Random}}  & Vanilla & 82.33  & 80.57  & 40.93  & ~ & 82.20  & 81.86  & 48.13  & ~ & 85.17  & 83.10  & 37.77  \\ 
        ~ & ~ & VCD & 87.23  & 86.19  & 42.43  & ~ & 84.37  & 83.80  & 46.50  & ~ & 87.17  & 85.65  & 39.43  \\ 
        ~ & ~ & M3ID & 86.53  & 85.18  & 40.87  & ~ & 79.77  & 78.66  & 44.83  & ~ & 85.97  & 83.99  & 37.63  \\ 
        ~ & ~ & Ritual & 85.80  & 83.81  & 37.73  & ~ & 85.33  & \textbf{83.76}  & 40.33  & ~ & 85.40  & 83.06  & 36.20  \\ 
        ~ & ~ & VaLiD & \textbf{89.03}  & \textbf{88.36}  & 44.23  & ~ & \textbf{85.90}  & 82.98  & 37.57  & ~ & \textbf{89.07}  & \textbf{88.14}  & 42.20  \\ \cline{2-14}
        ~ & \multirow{5}{*}{\textit{Popular}} & Vanilla & 81.10  & 79.48  & 42.10  & ~ & 79.10  & 79.30  & 50.97  & ~ & 84.73  & 82.70  & 38.27  \\ 
        ~ & ~ & VCD & 85.80  & 84.96  & 44.40  & ~ & 81.23  & 80.93  & 48.43  & ~ & 86.07  & 84.59  & 40.40  \\ 
        ~ & ~ & M3ID & 84.63  & 83.39  & 42.50  & ~ & 76.80  & 76.37  & 48.20  & ~ & 84.87  & 82.88  & 38.40  \\ 
        ~ & ~ & Ritual & 84.93  & 83.18  & 39.60  & ~ & 82.87  & 81.53  & 42.80  & ~ & 84.84  & 82.55  & 36.90  \\ 
        ~ & ~ & VaLiD & \textbf{87.17} & \textbf{86.65}  & 46.10  & ~ & \textbf{83.43}  & \textbf{81.59}  & 39.97  & ~ & \textbf{87.40}  & \textbf{86.51}  & 43.40  \\ \cline{2-14}
        ~ & \multirow{5}{*}{\textit{Adversarial}} & Vanilla & 78.63  & 77.76  & 46.10  & ~ & 76.87  & 77.58  & 53.20  & ~ & 83.33  & 81.63  & 40.73  \\ 
        ~ & ~ & VCD & 81.40  & 80.84  & 47.07  & ~ & 78.53  & 78.69  & 50.73  & ~ & 84.27  & 83.00  & 42.53  \\ 
        ~ & ~ & M3ID & 81.70  & 80.80  & 45.30  & ~ & 74.93  & 75.52  & 52.40  & ~ & 83.23  & 81.45  & 40.37  \\ 
        ~ & ~ & Ritual & 82.60  & 81.05  & 41.80  & ~ & 80.77  & \textbf{79.73}  & 44.90  & ~ & 82.80  & 80.45  & 38.00  \\ 
        ~ & ~ & VaLiD & \textbf{83.20}  & \textbf{83.29}  & 50.53  & ~ & \textbf{81.33}  & 79.59  & 41.47  & ~ & \textbf{84.27}  & \textbf{83.80}  & 47.13  \\ \hline\hline
        \multirow{15}{*}{\rotatebox{90}{\textbf{\normalsize A-OKVQA \cite{aokvqa}}}} & \multirow{5}{*}{\textit{Random}} & Vanilla & 85.53  & 85.11  & 47.13  & ~ & 80.50  & 81.49  & 55.37  & ~ & 86.80  & 85.58  & 41.53  \\ 
        ~ & ~ & VCD & 87.90  & 87.72  & 49.40  & ~ & 83.13  & 83.75  & 53.80  & ~ & 87.80  & 86.84  & 42.73  \\ 
        ~ & ~ & M3ID & 87.27  & 86.86  & 46.93  & ~ & 78.53  & 78.36  & 49.20  & ~ & 87.43  & 86.21  & 41.10  \\ 
        ~ & ~ & Ritual & 89.27  & 88.69  & 44.93  & ~ & 84.53  & 83.73  & 45.07  & ~ & 86.10  & 84.41  & 39.16  \\ 
        ~ & ~ & VaLiD & \textbf{90.53}  & \textbf{90.68}  & 51.60  & ~ & \textbf{87.73}  & \textbf{87.24}  & 46.13  & ~ & \textbf{89.36}  & \textbf{89.04}  & 47.03  \\ \cline{2-14}
        ~ & \multirow{5}{*}{\textit{Popular}} & Vanilla & 81.43  & 81.29  & 49.23  & ~ & 76.57  & 78.48  & 58.90  & ~ & 86.73  & 85.37  & 40.67  \\ 
        ~ & ~ & VCD & 82.47  & 83.03  & 53.33  & ~ & 78.76  & 80.36  & 58.10  & ~ & 87.53  & 86.66  & 43.47  \\ 
        ~ & ~ & M3ID & 83.53  & 83.48  & 49.67  & ~ & 76.67  & 77.08  & 51.80  & ~ & 87.27  & 86.09  & 41.53  \\ 
        ~ & ~ & Ritual & \textbf{86.60}  & 86.38  & 48.40  & ~ & 81.83  & 81.47  & 48.03  & ~ & 85.77  & 83.99  & 38.90  \\ 
        ~ & ~ & VaLiD & 86.17  & \textbf{86.97}  & 56.17  & ~ & \textbf{83.07}  & \textbf{83.13}  & 50.40  & ~ & \textbf{88.53}  & \textbf{88.21}  & 47.27  \\ \cline{2-14}
        ~ & \multirow{5}{*}{\textit{Adversarial}} & Vanilla & 75.47  & 76.81  & 55.80  & ~ & 70.43  & 74.34  & 65.23  & ~ & 80.87  & 80.17  & 46.47  \\
        ~ & ~ & VCD & 76.60  & 78.60  & 59.33  & ~ & 72.83  & 76.08  & 63.57  & ~ & 81.80  & 81.59  & 48.87  \\ 
        ~ & ~ & M3ID & 76.97  & 78.55  & 57.37  & ~ & 68.23  & 71.16  & 60.17  & ~ & 81.40  & 80.80  & 46.87  \\ 
        ~ & ~ & Ritual & \textbf{79.07}  & 79.81  & 53.67  & ~ & 74.63  & 75.79  & 54.77  & ~ & 80.83  & 79.66  & 44.23  \\ 
        ~ & ~ & VaLiD & 78.47  & \textbf{80.98}  & 60.60  & ~ & \textbf{76.83}  & \textbf{78.19}  & 56.23  & ~ & \textbf{82.33}  & \textbf{83.03}  & 54.13  \\ \hline\hline
        \multirow{15}{*}{\rotatebox{90}{\textbf{\normalsize GQA \cite{gqa}}}} & \multirow{5}{*}{\textit{Random}} & Vanilla & 84.70  & 83.97  & 45.43  & ~ & 79.17  & 80.47  & 56.70  & ~ & 84.40  & 83.27  & 43.27  \\ 
        ~ & ~ & VCD & 87.90  & 87.67  & 48.10  & ~ & 82.10  & 82.75  & 53.77  & ~ & 86.73  & 85.94  & 44.33  \\ 
        ~ & ~ & M3ID & 86.20  & 85.70  & 46.53  & ~ & 78.37  & 78.10  & 48.77  & ~ & 85.93  & 84.94  & 43.40  \\ 
        ~ & ~ & Ritual & 87.93  & 87.14  & 43.87  & ~ & 83.26  & 82.23  & 44.13  & ~ & 85.80  & 84.42  & 41.13  \\ 
        ~ & ~ & VaLiD & \textbf{89.73}  & \textbf{89.79}  & 50.53  & ~ & \textbf{86.53}  & \textbf{85.82}  & 45.07  & ~ & \textbf{90.07}  & \textbf{89.81}  & 47.53  \\ \cline{2-14}
        ~ & \multirow{5}{*}{\textit{Popular}} & Vanilla & 77.83  & 78.23  & 51.83  & ~ & 74.23  & 76.71  & 60.63  & ~ & 80.50  & 80.03  & 47.63  \\ 
        ~ & ~ & VCD & 80.07  & 81.10  & 55.47  & ~ & 76.63  & 78.60  & 59.17  & ~ & 82.33  & 82.09  & 48.67  \\ 
        ~ & ~ & M3ID & 78.60  & 78.84  & 51.13  & ~ & 74.07  & 74.92  & 53.40  & ~ & 81.07  & 80.29  & 46.07  \\ 
        ~ & ~ & Ritual & 82.08  & 82.58  & 48.73  & ~ & 79.33  & 79.29  & 49.80  & ~ & 82.56  & 81.26  & 43.03  \\ 
        ~ & ~ & VaLiD & \textbf{82.57}  & \textbf{83.82}  & 57.77  & ~ & \textbf{80.10}  & \textbf{80.38}  & 51.43  & ~ & \textbf{84.23}  & \textbf{84.66}  & 52.77  \\ \cline{2-14}
        ~ & \multirow{5}{*}{\textit{Adversarial}} & Vanilla & 75.70  & 76.57  & 53.70  & ~ & 71.13  & 74.69  & 64.07  & ~ & 79.33  & 79.17  & 49.20  \\
        ~ & ~ & VCD & 76.30  & 78.08  & 58.10  & ~ & 73.27  & 76.60  & 64.27  & ~ & 81.10  & 81.11  & 50.03  \\ 
        ~ & ~ & M3ID & 77.20  & 78.08  & 54.00  & ~ & 68.80  & 71.06  & 57.80  & ~ & 79.83  & 79.61  & 48.90  \\ 
        ~ & ~ & Ritual & \textbf{80.10}  & 80.47  & 51.90  & ~ & 74.67  & 75.75  & 54.47  & ~ & 80.33  & 79.31  & 45.06  \\ 
        ~ & ~ & VaLiD & 78.50 & \textbf{80.69} & 61.37 & ~ & \textbf{77.13} & \textbf{78.05} & 54.20 & ~ & \textbf{82.40} & \textbf{83.25} & 55.07 \\ \bottomrule
    \end{tabular}
    }
    \caption{
        \textbf{Results on POPE benchmark.} VaLiD consistently outperforms the contrastive decoding baseline: VCD \cite{decodingVCD}, M3ID \cite{decodingm3id} and Ritual \cite{decodingritual}. The best performances within each setting are \textbf{bolded}. VCD, M3ID and Ritual are reproduced within our evaluation setting.
    }
    \label{tab:pope}
    \vspace{-0.3cm}
\end{table*}

%% file: table_amber.tex
\begin{table*}[t!]
    \centering
    \renewcommand\arraystretch{1.08}
    \setlength{\tabcolsep}{10pt}
    \resizebox{0.9\linewidth}{!}{%
    \begin{tabular}{c c c c c c c c c c c c c}
    \toprule
        \multirow{2}{*}{\textbf{Model}} & \multirow{2}{*}{\textbf{Method}} & \multicolumn{2}{c}{\textbf{State}} & ~ & \multicolumn{2}{c}{\textbf{Action}} & ~ & \multicolumn{2}{c}{\textbf{Number}} & ~ & \multicolumn{2}{c}{\textbf{Relation}} \\ \cline{3-4}\cline{6-7}\cline{9-10}\cline{12-13}

        ~ & ~ & Acc.$\uparrow$ & F1$\uparrow$ & ~ & Acc.$\uparrow$ & F1$\uparrow$ & ~ & Acc.$\uparrow$ & F1$\uparrow$ & ~ & Acc.$\uparrow$ & F1$\uparrow$  \\ \hline
        
        \multirow{5}{*}{LLaVA-v1.5} & Vanilla & 69.77  & 69.20  & ~ & 78.91  & 78.00  & ~ & 71.81  & 70.47  & ~ & 58.59  & 53.91  \\ 
        ~ & VCD & 71.87  & 71.31  & ~ & 81.57  & 81.61  & ~ & 72.54  & 72.50  & ~ & 60.40  & 56.56  \\ 
        ~ & M3ID & 70.45  & 69.65  & ~ & 83.96  & 83.61  & ~ & 74.23  & 73.06  & ~ & 58.05  & 53.34  \\
        ~ & Ritual & 74.22  & 73.28  & ~ & 85.35  & 84.74  & ~ & \textbf{80.60}  & 79.06  & ~ & 54.99  & 39.35  \\ 
        ~ & VaLiD & \textbf{75.15}  & \textbf{75.59}  & ~ & \textbf{86.61}  & \textbf{87.10}  & ~ & 78.67  & \textbf{80.58}  & ~ & \textbf{71.39}  & \textbf{71.73}  \\ \cline{1-13}
        \multirow{5}{*}{InstructBLIP} & Vanilla & 69.37  & 69.80  & ~ & 73.86  & 75.73  & ~ & 69.64  & 73.02  & ~ & 58.11  & 61.68  \\ 
        ~ & VCD & 73.95  & \textbf{74.53}  & ~ & 77.53  & 79.30  & ~ & 73.89  & 77.49  & ~ & 60.40  & 65.04  \\ 
        ~ & M3ID & 64.82  & 62.51  & ~ & 77.15  & 75.90  & ~ & 65.97  & 68.98  & ~ & 58.29  & 62.81  \\ 
        ~ & Ritual & 69.50  & 64.71  & ~ & 80.18  & 78.24  & ~ & 74.23  & 76.84  & ~ & \textbf{67.37}  & \textbf{70.15}  \\ 
        ~ & VaLiD & \textbf{73.98}  & 72.36  & ~ & \textbf{83.33}  & \textbf{83.25}  & ~ & \textbf{82.77}  & \textbf{83.31}  & ~ & 63.52  & 61.51  \\ \cline{1-13}
        \multirow{5}{*}{Qwen-VL} & Vanilla & 75.90 & 72.60 & ~ & 83.08 & 81.18 & ~ & 82.38 & 80.92 & ~ & 53.73 & 39.28 \\ 
        ~ & VCD & 77.50 & 74.75 & ~ & 85.10 & 83.88 & ~ & 83.59 & 82.79 & ~ & 56.85 & 45.44 \\
        ~ & M3ID & 76.41 & 73.36 & ~ & 86.61 & 85.36 & ~ & 83.69 & 82.59 & ~ & 53.79 & 39.69 \\ 
        ~ & Ritual & 76.55 & 73.22 & ~ & 83.71 & 81.49 & ~ & 85.96 & 84.76 & ~ & 49.46 & 25.11 \\ 
        ~ & VaLiD & \textbf{79.64} & \textbf{79.35} & ~ & \textbf{87.59} & \textbf{87.45} & ~ & \textbf{86.51} & \textbf{85.87} & ~ & \textbf{68.21} & \textbf{68.49} \\ \bottomrule
    \end{tabular}
    }
    \caption{
        \textbf{Results on AMBER benchmark.} The AMBER dataset includes attribute-level hallucinations, which contains state attributes (e.g., color, shape), action attributes (e.g., actions performed by humans or animals in the image), and numerical attributes (typically noted when an object appears multiple times in the image). Additionally, it includes relation-level hallucinations, which refers to whether there is direct contact between two objects in an image.
    }
    \label{tab:amber}
    \vspace{-0.4cm}
\end{table*}

%% file: mme.tex
\begin{figure*}[t!]
    \centering
    \begin{minipage}{0.33\linewidth}
        \centering
        \centerline{\includegraphics[width=\textwidth]{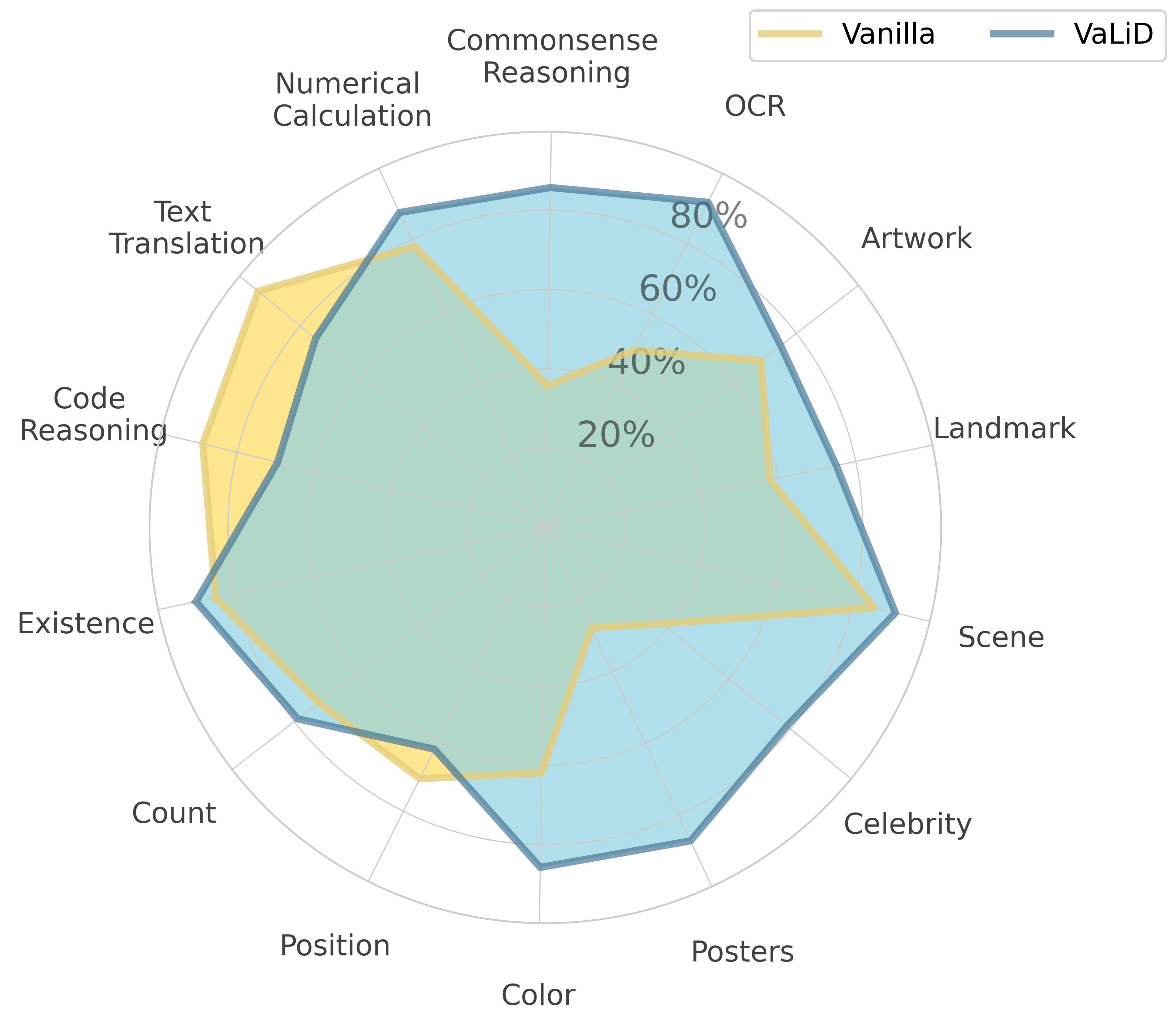}}
        \small\centerline{(a) LLaVA-v1.5}
    \end{minipage}
    %\qquad
    \begin{minipage}{0.33\linewidth}
        \centering
        \centerline{\includegraphics[width=\textwidth]{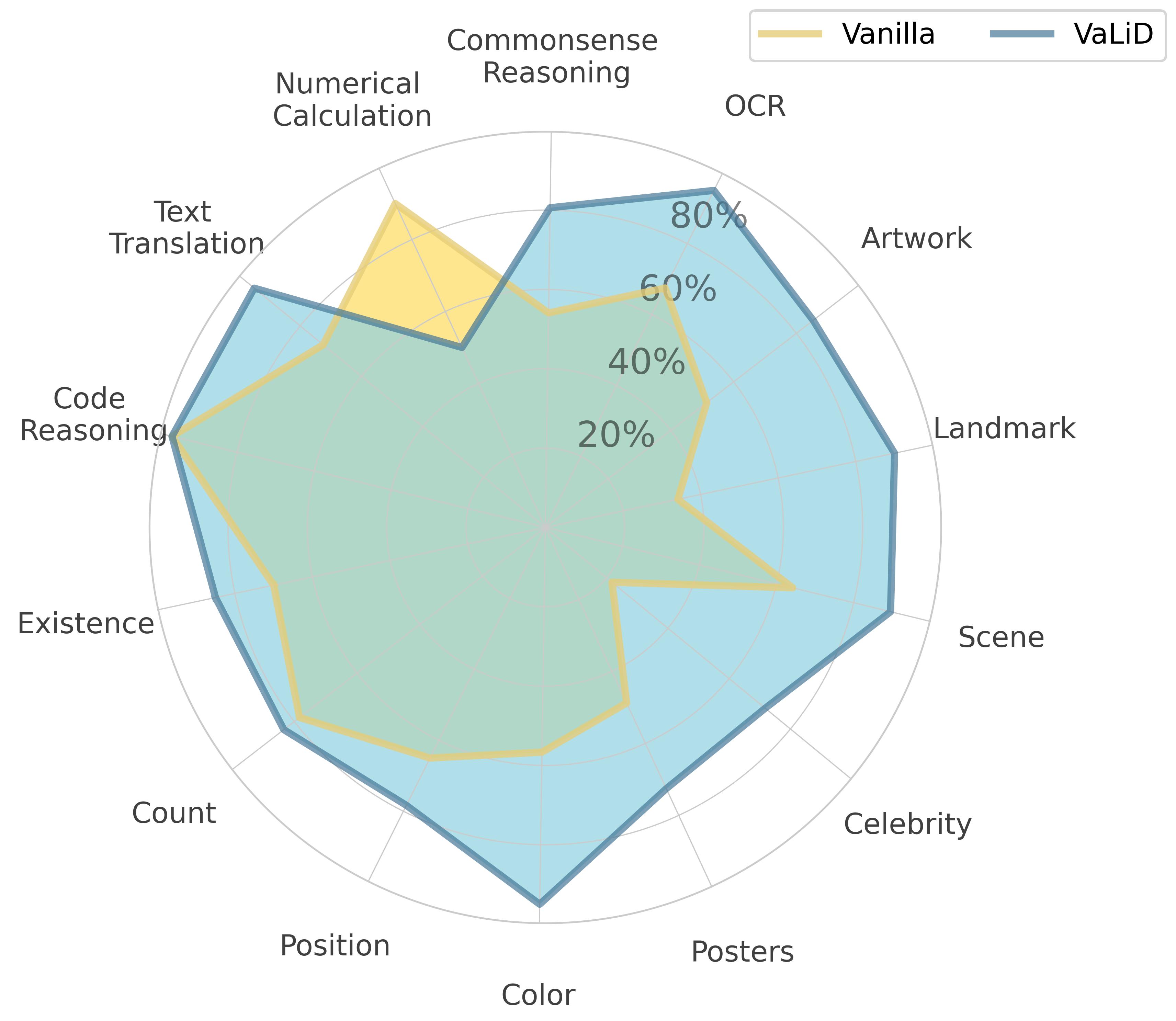}}
        \small\centerline{(b) InstructBLIP}
    \end{minipage}
    %\qquad
    \begin{minipage}{0.33\linewidth}
        \centering
        \centerline{\includegraphics[width=\textwidth]{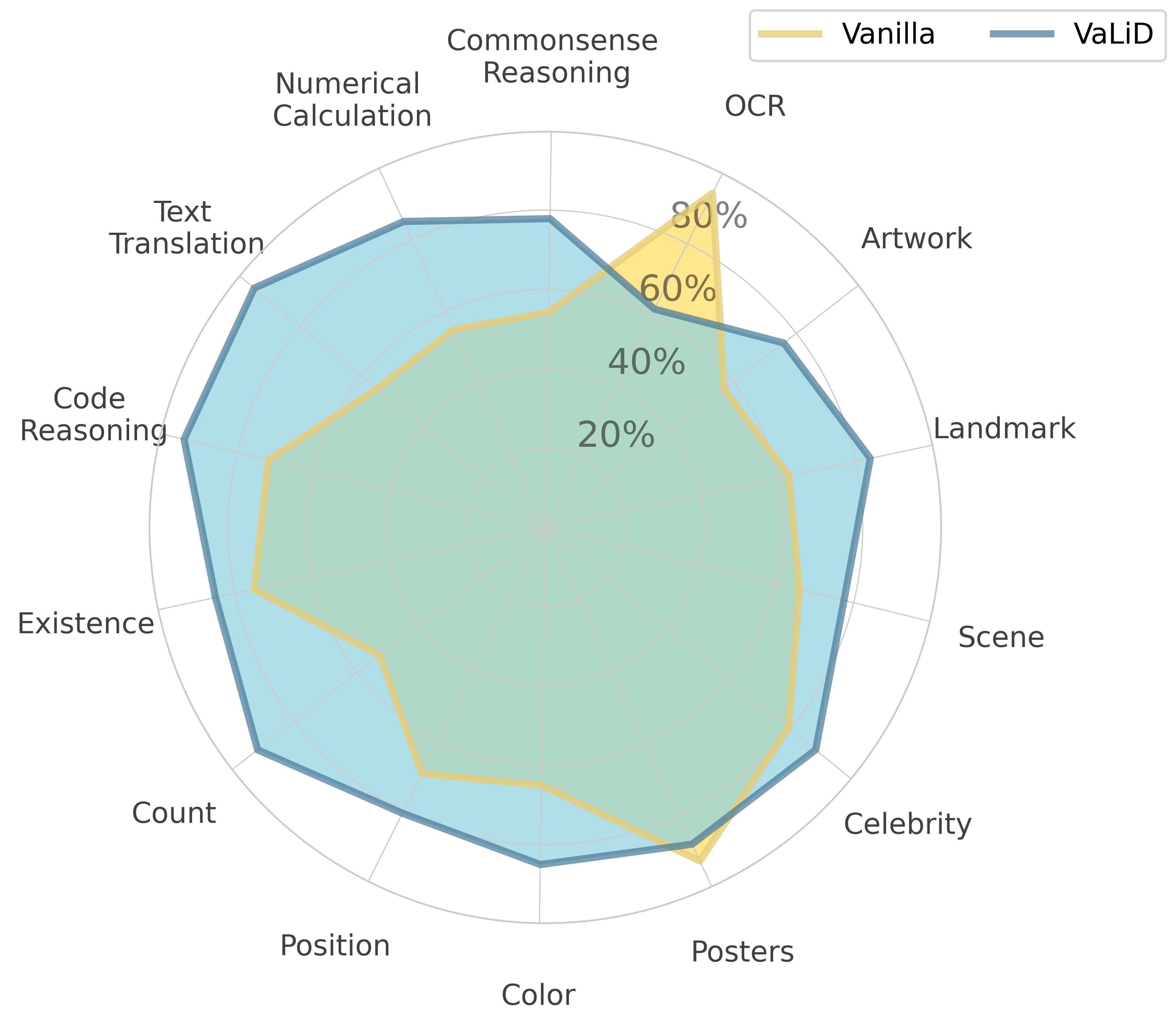}}
        \small\centerline{(c) Qwen-VL}
    \end{minipage}
    \caption{\textbf{Results of MME benchmark}. LVLMs with VaLiD achieve the best performance in \textbf{11} out of 14 categories for LLaVA-v1.5, \textbf{13} out of 14 categories for InstructBLIP, and \textbf{12} out of 14 categories for Qwen-VL. VaLiD not only mitigates hallucinations but also enhances the overall capabilities of LVLMs. Detailed results are provided in the supplementary material \ref{sec:appendix-mme}.}
    \label{fig:mme}%文中引用该图片代号
    \vspace{-0.4cm}
\end{figure*}

%% file: ablation_study.tex
\begin{table}[t!]
    \centering
    \renewcommand\arraystretch{1}
    \setlength{\tabcolsep}{5pt}
    \resizebox{1\linewidth}{!}{%
    \begin{tabular}{ccc|c>{\columncolor{gray!10}}c}
    \toprule
     & Random & Max entropy & Avg.-weighted & Entropy-weighted \\ \hline
        Acc.$\uparrow$ & 83.74	& 86.26	& 86.08	& \textbf{87.18} \\
        F1$\uparrow$   & 83.16	& 85.73	& 84.61	& \textbf{86.09} \\
    \bottomrule
    \end{tabular}
    }
    \vspace{-0.3cm}
    \caption{Ablation on layer selection on LLaVA-v1.5, MS-COCO.}
    \label{tab:rebuttal-tabble0}
    \vspace{-0.4cm}
\end{table}

%% file: table_final2.tex
\begin{table}[t!]
    \centering
    \renewcommand\arraystretch{1.2}
    \setlength{\tabcolsep}{5pt}
    \resizebox{1\linewidth}{!}{%
    \begin{tabular}{ccccccc}
    \toprule
        \multirow{2}{*}{\textbf{Method}} & \multicolumn{2}{c}{\textit{Random}} & \multicolumn{2}{c}{\textit{Popular}} & \multicolumn{2}{c}{\textit{Adversarial}} \\ \cline{2-7}
        ~ & Acc.$\uparrow$ & F1$\uparrow$ & Acc.$\uparrow$ & F1$\uparrow$ & Acc.$\uparrow$ & F1$\uparrow$ \\ \hline
        $\mathbb{P}_{\text{vcd}} \circ \mathbb{P}_{\text{valid}}$ & $89.00_{(1.09)}$ & $88.36_{(1.43)}$ & $86.50_{(0.66)}$ & $86.01_{(0.59)}$ & $83.07_{(0.54)}$ & $83.02_{(0.89)}$ \\ 
        $\mathbb{P}_{\text{valid}} \circ \mathbb{P}_{\text{vcd}}$ & $\textbf{89.40}_{(\textbf{1.49})}$ & $\textbf{88.59}_{(\textbf{1.66})}$ & $\textbf{87.23}_{(\textbf{1.39})}$ & $\textbf{86.50}_{(\textbf{1.08})}$ & $\textbf{83.93}_{(\textbf{1.40})}$ & $\textbf{83.49}_{(\textbf{1.36})}$ \\ \bottomrule
    \end{tabular}
    }
    \caption{Results of the POPE benchmark on LLaVA-v1.5. The values in brackets indicate the average improvement achieved by the method ensemble in the specified order, relative to VCD and VaLiD individually.}
    \label{tab:rebuttal-final2}
    \vspace{-0.5cm}
\end{table}

%% file: case_study.tex
\begin{figure*}[tp]
    \centering
    \includegraphics[width=1\linewidth]{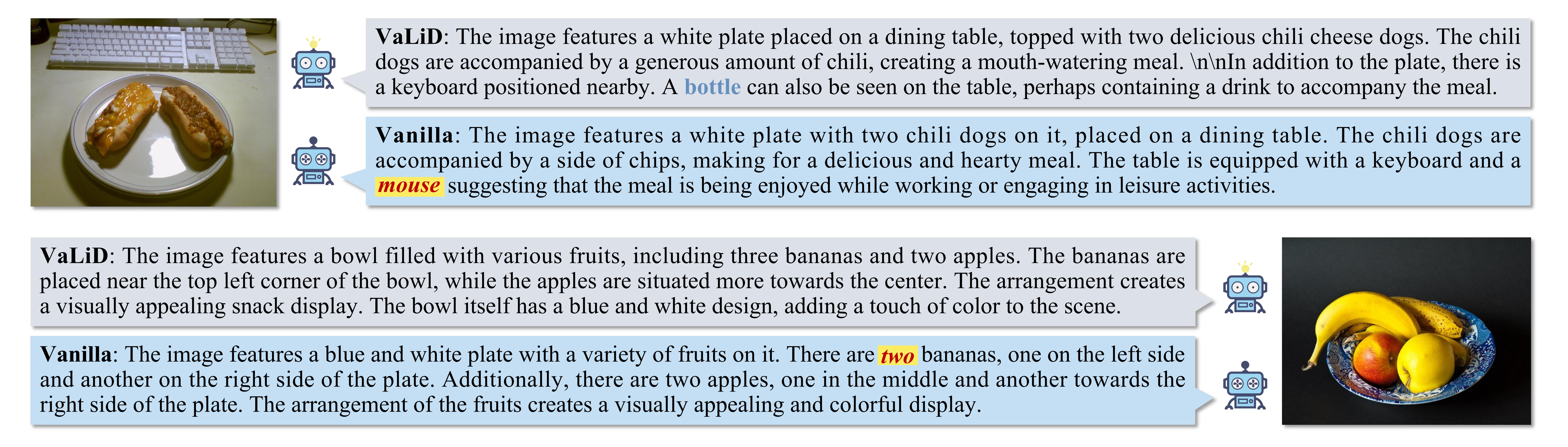}
    \vspace{-0.5cm}
        \caption{Illustration of hallucination correction by our proposed VaLiD with two examples from AMBER dataset \cite{AMBER}. Hallucinated objects from vanilla decoding are highlighted in \colorbox{yellow}{\textcolor{red}{\emph{\textbf{red}}}}.}
    \label{fig: case_study}
    \vspace{-0.35cm}
\end{figure*}

%% file: table_ritual_vcd_valid.tex
\begin{table}[t!]
    \centering
    \renewcommand\arraystretch{1}
    \setlength{\tabcolsep}{7pt}
    \resizebox{1\linewidth}{!}{%
    \begin{tabular}{ccccccc}
    \toprule
        \multirow{2}{*}{\textbf{Method}} & \multicolumn{2}{c}{\textit{Random}} & \multicolumn{2}{c}{\textit{Popular}} & \multicolumn{2}{c}{\textit{Adversarial}} \\ \cline{2-7}
        ~ & Acc.$\uparrow$ & F1$\uparrow$ & Acc.$\uparrow$ & F1$\uparrow$ & Acc.$\uparrow$ & F1$\uparrow$ \\ \hline
        $\mathbb{P}_{\text{vcd}} \circ \mathbb{P}_{\text{ritual}}$ & 86.23 & 84.37 & 85.77 & 84.18 & 82.70 & 81.23 \\ 
        $\mathbb{P}_{\text{valid}} \circ \mathbb{P}_{\text{ritual}}$ & \textbf{87.13} & \textbf{85.61} & \textbf{86.17} & \textbf{84.67} & \textbf{84.30} & \textbf{83.01} \\ \bottomrule
    \end{tabular}
    }
    \caption{Results of compatibility on LLaVA-v1.5.}
    \label{tab:rebuttal-tabble2}
    \vspace{-0.5cm}
\end{table}

%% file: rebuttal1.tex
% \vspace{-0.3cm}
\begin{table}[t!]
    \centering
    \renewcommand\arraystretch{1}
    \setlength{\tabcolsep}{10pt}
    \resizebox{1\linewidth}{!}{%
    \begin{tabular}{ccccccc}
    \toprule
        \multirow{2}{*}{\textbf{Method}} & \multicolumn{2}{c}{\textit{Random}} & \multicolumn{2}{c}{\textit{Popular}} & \multicolumn{2}{c}{\textit{Adversarial}} \\ \cline{2-7}
        ~ & Acc.$\uparrow$ & F1$\uparrow$ & Acc.$\uparrow$ & F1$\uparrow$ & Acc.$\uparrow$ & F1$\uparrow$ \\ \hline
        DOLA \cite{DOLA} & 87.40 & 88.57 & 80.47 & 83.32 & 69.53 & 76.21 \\
        DAMO \cite{damo} & 87.90 & 88.93 & 81.27 & 83.84 & 70.77 & 76.88 \\ 
        VaLiD & \textbf{89.93} & \textbf{89.85} & \textbf{86.03} & \textbf{86.47} & \textbf{79.13} & \textbf{81.13} \\
        \bottomrule
    \end{tabular}
    }
    \caption{Comparison with language-layer decoding approach.}
    \label{tab:rebuttal-tabble1}
    \vspace{-0.6cm}
\end{table}

%% file: X_suppl.tex
\clearpage
\appendix
\setcounter{page}{1}
\maketitlesupplementary

\section{Experiment Settings of Encoding Distortion}\label{EDR}
To quantitatively assess the distortion issue in the visual encoding process of LVLMs, we propose the \textbf{Encoding Distortion Rate (EDR)}. This metric quantifies the proportion of samples that are correctly decoded at early visual layers but incorrectly decoded at the standard output layer. Specifically, we decode the results for each layer by passing all hidden layer features of the visual encoder to the LLM. Considering that displaying the decoding accuracy of each layer separately may lead to the problem of sparse accuracy distribution, and there are certain similarities in the features between layers. To present the findings more intuitively, we summarize the accuracy statistics based on predefined layer groupings, referred to as ``buckets". In the case of LLaVA-v1.5, which comprises 25 layers, we organize the visual layers into the following buckets: $[0, 3]$, $[4, 7]$, $[8, 11]$, $[12, 15]$, $[16, 19]$, $[20, 24]$. A bucket is considered to yield correct answers if the visual features within it enable accurate decoding results by the LLMs. 

We evaluate the EDR score for three representative models (LLaVA-v1.5, InstructBLIP, and Qwen-VL) using the AMBER dataset. We reformulate each question from its original discriminative format in the AMBER dataset into a generative response task. Specifically, for tasks such as visual counting, we transform the question structure from a discriminative format (e.g., \textit{``Are there N [obj] in this image?"}) to a generative formulation (e.g., \textit{``How many [obj] are there in this image?"}). To facilitate the calculation of accuracy, we add the following instruction to the query: \textit{``Please generate the answer with one word."} This formulation enables the model to generate a direct count of the specified objects, thereby offering a more precise evaluation of its ability to interpret and reason about visual content.

\section{Details of Other Baseline Methods}\label{baseline-settings}
In this work, we compare VaLiD with VCD \cite{decodingVCD}, M3ID \cite{decodingm3id} and Ritual \cite{decodingritual}. The method and implementation details for these approaches are provided below: 

\noindent\textbf{VCD} attributes the hallucination of LVLMs to language priors and statistical biases. Specifically, given a visual input $v$ and a textual query $x$, VCD first adds Gaussian noise to the image $v$, resulting in $v'$. Then a new contrastive probability distribution is computed as follows:
$$ P_{vcd} = (1+\alpha) P_{\theta} (y_{t} \vert v,x,y_{<t}) - \alpha P_{\theta} (y_{t} \vert v',x,y_{<t}) $$
\noindent where $y_{<t}$ denotes the generated tokens. In our implementation, we follow the default setting and set $\alpha=1$.

\vspace{0.1cm}
\noindent\textbf{M3ID} contrasts output distributions from original visual inputs and pure text inputs without visual information. The contrastive probability distribution is computed as follows:
$$ P_{m3id} = (1+\dfrac{1-e^{-\lambda t}}{e^{-\lambda t}})P_{\theta} (y_{t} \vert v,x,y_{<t}) - \dfrac{1-e^{-\lambda t}}{e^{-\lambda t}} P_{\theta} (y_{t} \vert x,y_{<t}) $$
\noindent where we set the hyperparameter $\lambda=0.02$.

\vspace{0.1cm}
\noindent\textbf{Ritual} applies common image transformations (e.g., crop, flip, color jitter,
etc.) to the original visual input $v$, resulting in a transformed version $v^{T}$. Both original and transformed images are used to generate the response. The probability distribution is calculated as follows:
$$ P_{ritual} = P_{\theta} (y_{t} \vert v,x,y_{<t}) + \alpha P_{\theta} (y_{t} \vert v,x,y_{<t}) $$
\noindent where we follow the default setting and set $\alpha=3$.

\section{Adaptive Reliability Constraint}\label{adaptive-reliability-constraint}
In order to prevent the newly generated distribution mistakenly penalizing the valid outputs, leading to decoding results that do not adhere to basic language standards and commonsense reasoning, we follow \cite{2.1_1} and adapt an adaptive reliability constraint to avoid unexpected penalty, as shown in Eq. \ref{eq:4}.

\vspace{-0.25cm}
\begin{equation}
    \label{eq:4}
    \begin{gathered}
        \begin{aligned}
            & \mathcal{V}_{\text {head }}\left(y_{<t}\right)= 
            \{y_t \in \mathcal{V}\vert  \\
            &P_{\theta}\left(y_t \mid v,x,y_{<t}\right) \geq \beta \max _w P_{\theta}\left(w \mid v,x,y_{<t}\right)\},
        \end{aligned}
        \\
        \begin{aligned}
            P_{valid}\left(y_t \mid v, x, y_{< t} \right) = 0, \text{ if } y_t \notin \mathcal{V}_{\text {head }}\left(y_{<t}\right),
        \end{aligned}
    \end{gathered}
\end{equation}

\noindent where $\beta$ is a hyper parameter, $P_{valid}$ denotes the token distribution after VaLiD's decoding. If $y_{t}$ is not in $\mathcal{V}_{\text{head}}(y_{<t})$, it indicates that the token has a very low probability under the original distribution, suggesting it is likely an unreasonable prediction. Therefore, we set the probability of the same position $y_{t}$ in the corrected distribution to zero, preventing the score of these $y_{t}$ tokens from being abnormally amplified after contrastive adjustment. The ablation study of $\beta$ is available in \ref{aba-study}.

\input{table_mme}

\input{layer_selection}

\section{Experimental Settings}\label{exp-settings}
\noindent\textbf{Layer Selection of LVLMs}: 
We construct a validation set with 300 question-answer (Q, A) pairs from MS-COCO dataset according to the settings of POPE \cite{pope}. This validation set is carefully designed to include three distinct data categories - random, popular, and adversarial - with 100 samples each, ensuring comprehensive coverage of different evaluation scenarios. Before conducting the main experiments, we use the validation set to determine the best candidate set of different LVLMs. Specifically, the visual hidden layer of an LVLM will be divided into distinct buckets, upon which we perform VaLiD validation to determine the most effective candidate layer set. 

To ensure the computational efficiency, our analysis is restricted to odd-indexed layers (i.e., 1st, 3rd, etc.). For instance, in the case of LLaVA-v1.5, which contains 25 visual hidden layers, we initially divide the layers into two distinct buckets: $[1, 3, 5, 7, 9, 11]$ and $[13, 15, 17, 19, 21, 23]$. Table \ref{tab:layer-selection} shows the results of the validation set, we finally identify $[13, 15, 17, 19, 21, 23]$ as the optimal candidate set for the 25-layer LLaVA-v1.5-7B. For the 39-layer InstructBLIP-7B, we choose layers $[27, 29, 31, 33, 35, 37]$. For the 48-layer Qwen-VL-7B, we use layers $[37, 39, 41, 43, 45, 47]$. 

\vspace{0.15cm}

\noindent\textbf{Metrics}: We follow the official implementations for the POPE, MME, and AMBER benchmarks. Specifically, hallucination evaluation on these benchmarks is generally defined as a binary classification task. We utilize multiple metrics, including Accuracy, Precision, Recall, and F1-score.

\section{Additional Numerical Results on POPE}\label{sec:appendix-pope_score}
Table \ref{tab:appendix-pope_llava}, Table \ref{tab:appendix-pope_blip2} and Table \ref{tab:appendix-pope_qwen} provide the detail numerical results of POPE, including accuracy, precision, recall, f1-score. Additionally, we report the \textit{Yes} ratio in POPE to further showcase LVLMs' behavior under different sampling difficulty scenarios, where \textit{Yes} ratio denotes the proportion of answering ``Yes" to the given question. The results demonstrate that, as the sampling difficulty increases, the model is more inclined to answer Yes under the VaLiD decoding strategy.

\section{Numerical Results of VaLiD on MME}\label{sec:appendix-mme}
In this section, we evaluate the impact of various decoding strategies on the model's original performance while mitigating hallucinations, using the MME dataset. As shown in Table \ref{sec:table-mme}, VaLiD consistently outperforms the vanilla model across most tasks, particularly in reasoning tasks such as numerical calculation and text translation. Furthermore, in most cases, VaLiD surpasses other methods (VCD \cite{decodingVCD}, M3ID \cite{decodingm3id}, and Ritual \cite{decodingritual}), highlighting its superior ability to reduce hallucinations and improve the reliability of the generated outputs. (1) Our MME-hallu results outperform the baseline methods, consistent with the results on POPE and AMBER datasets. (2) Performance on MME non-hallu sets almost surpass vanilla method, aligning with our expectations that VaLiD preserves LVLMs' original capabilities.

\input{table_vcd_valid}

\section{Experimental Settings of Ablation Studies}\label{aba-study}
We conduct ablation experiments on (i) layer selection and (ii) fusion strategy to justify the effectiveness of VaLiD. Regarding the selection strategy for visual layers, we first investigate a randomized approach where we sample a layer uniformly from the visual encoder. To ensure statistical robustness, we repeat this process three times with distinct random seeds and compute the average of the final results. This random selection baseline is then compared with our proposed entropy-based layer selection methodology. In our entropy-based approach, rather than exhaustively searching through all visual layers to identify the maximum entropy layer - which would be computationally intensive - we select the layer exhibiting the highest entropy within the optimal subset. 

Regarding the distribution fusion strategy, we compare two distinct approaches: (1) average-weighted fusion and (2) entropy-weighted fusion. These strategies are specifically applied to the visual layers that correspond to the top-k highest entropy values. The average-weighted fusion assigns equal importance to each selected layer, while the entropy-weighted fusion dynamically adjusts the contribution of each layer based on its relative entropy value, thereby giving greater weight to layers with higher information content.

We also report the ablation study on the selection of hyper parameter $\alpha$ and $\beta$, where $\alpha$ determines the relative influence of original and reference distribution and $\beta$ is related to the adaptive reliability constraint. We evaluate LLaVA-v1.5 on the validation set constructed in \ref{exp-settings}, as shown in Table \ref{tab:abl-hyper} and Table \ref{tab:abl-hyper2}. The results indicate that our default choice of $\alpha=3$ and $\beta=0.1$ yields the best performance.

\input{abl-hyperpara}

\section{VaLiD is More Compatible than VCD}\label{sec:appendix-compatible}

\begin{claim}\label{claim}
    Let $\mathbb{P}_{\text{valid}} \circ \mathbb{P}_{\text{vcd}}$ and $\mathbb{P}_{\text{vcd}} \circ \mathbb{P}_{\text{valid}}$ represent the application of VCD and VaLiD in different sequences. Assuming truncation effects are negligible, then we have 
    $$ \Vert \mathbb{P}_{\text{valid}} \circ \mathbb{P}_{\text{vcd}} - \mathbb{P}_{\text{vcd}} \circ \mathbb{P}_{\text{valid}} \Vert = \mathcal{K} \vert P_{\text{noi}} - P_{\text{ent}} \vert $$
    where $\mathbb{P}$ indicate an operator formed as $(1+\alpha)P_{\text{ori}} - \alpha P_{\text{ref}}$, $\circ$ represents the composition operator, $P_{\text{noi}}$ and $P_{\text{ent}}$ denote the reference distribution construct by VCD and VaLiD, respectively, $\mathcal{K}$ denotes a constant value.
\end{claim}

\input{abl-hyperpara2}

\input{sup-case_study}

\begin{proof}\label{proof}
    \noindent Let $\mathbb{P}_{\text{valid}} \circ \mathbb{P}_{\text{vcd}}$ and $\mathbb{P}_{\text{vcd}} \circ \mathbb{P}_{\text{valid}}$ represent the application of VCD and VaLiD in different orders. Considering that $\mathbb{P}$ indicate an operator formed as $(1+\alpha)P_{\text{ori}} - \alpha P_{\text{ref}}$, where $\alpha$ and $\alpha^{'}$ are two positive coefficients to control the contrast intensity for VCD and VaLiD, respectively. Assuming truncation effects are negligible, then we have 
\begin{flalign*}
    &\mathbb{P}_{\text{valid}} \cdot \mathbb{P}_{\text{vcd}} (P_{\text{ori}}) = \mathbb{P}_{\text{valid}} \left[ (1 + \alpha) P_{\text{ori}} - \alpha P_{\text{noi}} \right] &\\
    &= (1 + \alpha^{'}) \cdot (1 + \alpha) P_{\text{ori}} - \alpha^{'} P_{\text{noi}} - \alpha^{'} P_{\text{ent}} &\\
    &= (1 + \alpha^{'}) (1 + \alpha) P_{\text{ori}} - \left[ \alpha (1 + \alpha^{'}) P_{\text{noi}} + \alpha^{'} P_{\text{ent}} \right] &\\
    &\mathbb{P}_{\text{vcd}} \cdot \mathbb{P}_{\text{valid}} (P_{\text{ori}}) = \mathbb{P}_{\text{vcd}} \left[ (1 + \alpha^{'}) P_{\text{ori}} - \alpha^{'} P_{\text{ent}} \right] &\\
    &= (1 + \alpha) \left[ (1 + \alpha^{'}) P_{\text{ori}} - \alpha^{'} P_{\text{ent}} \right] - \alpha \cdot P_{\text{noi}} &\\
    &= (1 + \alpha) (1 + \alpha^{'}) P_{\text{ori}} - \left[ (1 + \alpha) \alpha^{'} P_{\text{ent}} + \alpha P_{\text{noi}} \right] &\\
    &\Vert\mathbb{P}_{\text{valid}} \cdot \mathbb{P}_{\text{vcd}}  - \mathbb{P}_{\text{vcd}} \cdot \mathbb{P}_{\text{valid}} \Vert &\\
    &= \vert - \left[ \alpha (1 + \alpha^{'}) P_{\text{noi}} + \alpha^{'} P_{\text{ent}} \right] + \left[ (1 + \alpha) \alpha^{'} P_{\text{ent}} + \alpha P_{\text{noi}} \right] \vert  &\\
    &=\vert \alpha^{'} P_{\text{ent}} + \alpha\alpha^{'} P_{\text{ent}} + \alpha P_{\text{noi}} - \alpha P_{\text{noi}} - \alpha\alpha^{'} P_{\text{noi}} - \alpha^{'} P_{\text{ent}} \vert &\\
    &=\alpha\alpha^{'} \vert P_{\text{ent}} - P_{\text{noi}}\vert &\\
    &=\mathcal{K} \vert P_{\text{ent}} - P_{\text{noi}}\vert &
\end{flalign*}
where $P_{\text{ori}}$, $P_{\text{noi}}$ and $P_{\text{ent}}$ denote the original distribution, the reference distribution constructed by VCD, and the reference distribution constructed by VaLiD, respectively. $\mathcal{K}$ denotes a constant value, determined by $\alpha$ and $\alpha^{'}$.
\end{proof}

\noindent The results show that the order in which VCD and VaLiD are applied affects the model's performance. As stated in Claim \ref{claim}, the variation in probability distributions resulting from different application orders is proportional to the difference in reference distributions constructed by the two methods. Combining the experimental results from the Table \ref{tab:vcd-valid}, we observe that applying VCD first, followed by VaLiD, leads to further performance improvements. This reinforces the importance of addressing distortions in the visual encoding process as a critical factor in mitigating model hallucinations.

\section{Some Case Studies}\label{sec:appendix-more_case}
We provide more case studies on all benchmarks (POPE \cite{pope}, AMBER \cite{AMBER}) to verify the efficacy of VaLiD in generative tasks. Figure \ref{fig: sup-case_study} presents some case studies that illustrate how vanilla decoding results in object hallucinations, while VaLiD generates responses more closely aligned with the visual content. In these examples, a bicycle is often co-occurring with people, which leads the model to produce hallucinated objects. In contrast, VaLiD effectively mitigates these issues. Additionally, VaLiD successfully identifies the number of object, a common issue for most LVLMs.

\input{pope-llava}

\input{pope-blip2}

\input{pope-qwen}

%% file: table_mme.tex
\begin{table*}[!t]
    \centering
    \renewcommand\arraystretch{1.2}
    \setlength{\tabcolsep}{4pt}
    \resizebox{1\linewidth}{!}{%
    \begin{tabular}{cccccccccccccccc}
    \toprule
        \multirow{2}{*}{\textbf{Model}} & \multirow{2}{*}{\textbf{Method}} & \multicolumn{10}{c}{\textbf{Perception}} & \multicolumn{4}{c}{\textbf{Cognition}} \\ \cline{3-16}
        ~ & ~ & existence & count & position & color & posters & celebrity & scene & landmark & artwork & OCR & \makecell{commonsense\\reasoning} & \makecell{numerical\\calculation} & \makecell{text\\translation} & \makecell{code\\reasoning} \\ \hline
        \multirow{5}{*}{\textbf{LLaVA-v1.5}} & Vanilla & 185.00  & 143.33  & \textbf{128.33}  & 143.33  & 113.95  & 109.12  & 151.25  & 134.75  & 113.75  & 115.00  & 100.71  & 62.50  & 75.00  & 80.00  \\ \cline{2-16}
        ~ & VCD \cite{decodingVCD} & 185.00  & 142.67  & 118.33  & 158.33  & 130.95  & \textbf{124.71}  & 149.75  & 138.00  & 109.75  & 110.00  & 116.43  & 87.50  & 47.50  & 70.00  \\ 
        ~ & M3ID \cite{decodingm3id} & 185.00  & 130.00  & 118.33  & 145.00  & 126.53  & 124.12  & 147.00  & 141.75  & 108.00  & 110.00  & 107.14  & 52.50  & 82.50  & \textbf{95.00}  \\ 
        ~ & Ritual \cite{decodingritual} & \textbf{190.00}  & 138.33  & 118.33  & \textbf{160.00}  & \textbf{141.50}  & 116.18  & \textbf{158.50}  & \textbf{150.00}  & \textbf{120.75}  & 115.00  & \textbf{116.43}  & 42.50  & \textbf{87.50}  & 85.00 \\ \cline{2-16}
        ~ & VaLiD & \textbf{190.00}  & \textbf{147.33}  & 125.00  & \textbf{160.00}  & \textbf{143.54}  & \textbf{129.41}  & \textbf{154.50}  & \textbf{145.00}  & \textbf{115.00}  & \textbf{127.50}  & \textbf{115.71}  & \textbf{70.00}  & 60.00  & 62.50  \\ \midrule\midrule
        
        \multirow{5}{*}{\textbf{InstructBLIP}} & Vanilla & 170.00  & 78.33  & 65.00  & 106.67  & 114.63  & 110.88  & 138.50  & 120.50  & 86.00  & 67.50  & 87.86  & \textbf{90.00}  & 72.50  & 77.50  \\ \cline{2-16}
        ~ & VCD \cite{decodingVCD} & 180.00  & \textbf{88.33}  & \textbf{75.00}  & \textbf{136.67}  & 120.41  & 120.00  & 150.00  & 138.75  & 96.00  & 80.00  & 97.14  & 60.00  & 62.50  & 47.50  \\
        ~ & M3ID \cite{decodingm3id} & 175.00  & 80.00  & 76.67  & 105.00  & 131.29  & 113.23  & 149.00  & 118.75  & 100.75  & 70.00  & \textbf{114.29}  & 67.50  & 77.50  & 70.00  \\
        ~ & Ritual \cite{decodingritual} & \textbf{180.00}  & 68.33  & 68.33  & 120.00  & \textbf{133.33}  & \textbf{152.65}  & \textbf{161.25}  & \textbf{149.25}  & \textbf{104.75}  & \textbf{140.00}  & 113.57  & 72.50  & \textbf{82.50}  & 60.00  \\ \cline{2-16}
        ~ & VaLiD & \textbf{185.00}  & \textbf{83.33}  & \textbf{78.33}  & \textbf{145.00}  & \textbf{121.77}  & \textbf{135.88}  & \textbf{153.75}  & \textbf{154.00}  & \textbf{109.75}  & \textbf{95.00}  & \textbf{106.43}  & 50.00  & \textbf{95.00}  & \textbf{77.50}  \\ \hline

        \multirow{5}{*}{\textbf{Qwen-VL}} & Vanilla & 175.00  & 131.67  & 141.67  & 165.00  & \textbf{174.15}  & 147.65  & 152.75  & 150.25  & 128.50  & 122.50  & 104.29  & 27.50  & 105.00  & 57.50  \\ \cline{2-16}
        ~ & VCD \cite{decodingVCD} & 170.00  & 131.67  & 151.67  & 175.00  & 167.35  & \textbf{156.47}  & 153.50  & 153.50  & \textbf{143.00}  & 110.00  & \textbf{112.14}  & 35.00  & 112.50  & 47.50  \\ 
        ~ & M3ID \cite{decodingm3id} & \textbf{180.00}  & \textbf{141.67}  & 145.00  & 180.00  & 168.37  & 152.35  & 151.50  & 157.50  & 131.75  & 125.00  & 101.43  & 42.50  & 107.50  & \textbf{67.50}  \\ 
        ~ & Ritual \cite{decodingritual} & 170.00  & 128.33  & \textbf{168.33}  & \textbf{180.00}  & 171.43  & 149.12  & \textbf{154.25}  & \textbf{169.25}  & 138.75  & \textbf{130.00}  & 104.29  & \textbf{60.00}  & \textbf{140.00}  & 57.50  \\ \cline{2-16}
        ~ & VaLiD & \textbf{185.00}  & \textbf{155.00}  & \textbf{148.33}  & \textbf{185.00}  & 170.41  & \textbf{152.94}  & \textbf{162.00}  & \textbf{167.00}  & \textbf{138.00}  & 80.00  & \textbf{127.86}  & \textbf{42.50}  & \textbf{145.00}  & \textbf{75.00} \\ \bottomrule
    \end{tabular}
    }
    \caption{\textbf{Numerical results of MME benchmark}. The table presents the MME scores of various models across multiple perception and cognition tasks. The left four columns represent the results on MME-Hallu set, while others reflects the performance of LVLM on multimodal understanding. The best performances and second best performances within each setting are \textbf{bolded}.}
    \label{sec:table-mme}
    \vspace{-0.4cm}
\end{table*}

%% file: layer_selection.tex
% \vspace{-0.3cm}
\begin{table}[t!]
    \centering
    \renewcommand\arraystretch{1.14}
    \setlength{\tabcolsep}{5pt}
    \resizebox{1\linewidth}{!}{%
    \begin{tabular}{cccccccc}
    \toprule
        \multirow{2}{*}{\textbf{Model}} & \multirow{2}{*}{\textbf{Bucket}} & \multicolumn{2}{c}{\textit{Random}} & \multicolumn{2}{c}{\textit{Popular}} & \multicolumn{2}{c}{\textit{Adversarial}} \\ \cline{3-8}
        
        ~ & ~ & Acc. & F1 & Acc. & F1 & Acc. & F1 \\ \hline
        \multirow{2}{*}{\textbf{LLaVA-v1.5}} & $[1,3,5,7,9,11]$ & 88.00  & 86.67  & 83.11  & 82.61  & 84.00  & 82.61  \\
        ~ & $[13,15,17,19,21,23]$ & \textbf{90.00}  &\textbf{ 89.13}  & \textbf{84.00}  & \textbf{82.98}  & \textbf{86.00}  & \textbf{85.11}  \\ \hline
        \multirow{3}{*}{\textbf{InstructBLIP}} & $[3,5,7,9,11,13]$ & 84.00  & 88.92  & 81.00  & 79.94  & 79.00  & 77.78  \\ 
        ~ & $[15,17,19,21,23,25]$ & 85.00  & 90.21  & 83.00  & 81.20  & \textbf{82.00}  & 78.02  \\ 
        ~ & $[27,29,31,33,35,37]$ & \textbf{87.00}  & \textbf{91.21} & \textbf{86.00}  & \textbf{83.76}  & 80.00  & \textbf{78.91}  \\ \hline
        \multirow{3}{*}{\textbf{Qwen-VL}} & $[13, 15, 17, 19, 21, 23]$ & 87.00  & 85.39  & 82.00  & 79.55  & 81.00  & 78.17  \\ 
        ~ & $[25, 27, 29, 31, 33, 35]$ & 88.00  & 86.67  & 84.00  & 82.22  & \textbf{82.00}  & \textbf{79.54}  \\
        ~ & $[37, 39, 41, 43, 45, 47]$ & \textbf{92.00}  & \textbf{91.49}  & \textbf{84.00}  & \textbf{82.61}  & 80.00  & 77.78 \\
        
        \bottomrule
    \end{tabular}
    }
    \caption{Layer selection of LVLMs using validation set.}
    \label{tab:layer-selection}
    \vspace{-0.5cm}
\end{table}

%% file: table_vcd_valid.tex
\begin{table*}[t!]
    \centering
    \renewcommand\arraystretch{1.15}
    \setlength{\tabcolsep}{7pt}
    \resizebox{1\linewidth}{!}{%
    \begin{tabular}{ccccccccccc}
    \toprule
        \multirow{2}{*}{\textbf{Type}} & \multirow{2}{*}{\textbf{Method}} & \multicolumn{3}{c}{\textbf{MSCOCO}} & \multicolumn{3}{c}{\textbf{A-OKVQA}} & \multicolumn{3}{c}{\textbf{GQA}} \\ \cline{3-11}
        ~ & ~ & Acc.$\uparrow$ & F1$\uparrow$ & Yes & Acc.$\uparrow$ & F1$\uparrow$ & Yes & Acc.$\uparrow$ & F1$\uparrow$ & Yes \\ \hline
        \multirow{2}{*}{\textit{Random}} & $\mathbb{P}_{\text{vcd}} \circ \mathbb{P}_{\text{valid}}$ & $89.00_{(1.09)}$ & $88.36_{(1.43)}$ & $44.47$ & $\textbf{90.43}_{(\textbf{1.52})}$  & $\textbf{90.61}_{(\textbf{1.83})}$ & $51.90$ & $\textbf{89.00}_{(\textbf{0.70})}$ & $\textbf{89.10}_{(\textbf{0.95})}$ & $50.93$ \\ 
        ~ &$\mathbb{P}_{\text{valid}} \circ \mathbb{P}_{\text{vcd}}$ & $\textbf{89.40}_{(\textbf{1.49})}$ & $\textbf{88.59}_{(\textbf{1.66})}$ & $42.93$ & $88.93_{(0.02)}$ & $88.06_{(-0.73)}$ & $42.67$ & $88.70_{(0.40)}$ & $87.83_{(-0.32)}$ & $42.83$ \\ \hline
        \multirow{2}{*}{\textit{Popular}} & $\mathbb{P}_{\text{vcd}} \circ \mathbb{P}_{\text{valid}}$ & $86.50_{(0.66)}$ & $86.01_{(0.59)}$ & $46.50$ & $85.87_{(1.62)}$ & $\textbf{86.73}_{(\textbf{1.98})}$ & $56.47$ & $82.33_{(1.20)}$ & $83.60_{(1.53)}$ & $57.73$ \\
        ~ &$\mathbb{P}_{\text{valid}} \circ \mathbb{P}_{\text{vcd}}$ & $\textbf{87.23}_{(\textbf{1.39})}$ & $\textbf{86.50}_{(\textbf{1.08})}$ & $44.57$ & $\textbf{86.90}_{(\textbf{2.65})}$ & $86.17_{(1.42)}$ & $44.70$ & $\textbf{87.67}_{(\textbf{6.54})}$ & $\textbf{86.98}_{(\textbf{4.91})}$ & $44.73$ \\ \hline
        \multirow{2}{*}{\textit{Adversarial}} & $\mathbb{P}_{\text{vcd}} \circ \mathbb{P}_{\text{valid}}$ & $83.07_{(0.54)}$ & $83.02_{(0.89)}$ & $49.73$ & $77.90_{(0.04)}$ & $80.61_{(0.75)}$ & $63.97$ & $77.17_{(-0.40)}$ & $79.52_{(0.22)}$ & $61.50$ \\ 
        ~ &$\mathbb{P}_{\text{valid}} \circ \mathbb{P}_{\text{vcd}}$ & $\textbf{83.93}_{(\textbf{1.40})}$ & $\textbf{83.49}_{(\textbf{1.36})}$ & $47.33$ & $\textbf{83.80}_{(\textbf{5.94})}$ & $\textbf{83.45}_{(\textbf{3.59})}$ & $47.87$ & $\textbf{83.93}_{(\textbf{6.37})}$ & $\textbf{83.56}_{(\textbf{4.26})}$ & $47.73$ \\ \bottomrule
    \end{tabular}
    }
    \caption{Results of the POPE benchmark on LLaVA-v1.5. The values in brackets indicate the average improvement achieved by the method ensemble in the specified order, relative to VCD and VaLiD individually.}
    \label{tab:vcd-valid}
    \vspace{-0.4cm}
\end{table*}

%% file: abl-hyperpara.tex
\begin{table}[!t]
    \centering
    \renewcommand\arraystretch{1}
    \setlength{\tabcolsep}{13pt}
    \resizebox{1\linewidth}{!}{%
    \begin{tabular}{c c c c c}
    \toprule
        Value & Accuracy & Precision & Recall & F1 \\ \hline
        $\alpha=0$ & 82.67  & 93.22  & 71.43  & 80.88  \\
        $\alpha=1.0$ & 86.00  & 93.08  & 78.57  & 85.21  \\
        $\alpha=2.0$ & 85.33  & 94.35  & 75.97  & 84.17  \\
        $\alpha=3.0$ & \textbf{88.00}  & 95.38  & 80.52  & \textbf{87.32} \\
        $\alpha=4.0$ & 87.67  & 95.35  & 79.87  & 86.93 \\
    \bottomrule
    \end{tabular}
    }
    \vspace{-0.2cm}
    \caption{Ablation on hyper parameter $\alpha$ selection.}
    \label{tab:abl-hyper}
    \vspace{-0.5cm}
\end{table}

%% file: abl-hyperpara2.tex
\begin{table}[!t]
    \centering
    \renewcommand\arraystretch{1}
    \setlength{\tabcolsep}{15pt}
    \resizebox{1\linewidth}{!}{%
    \begin{tabular}{c c c c c}
    \toprule
        Value & Accuracy & Precision & Recall & F1 \\ \hline
        $\beta=0$ & 87.67  & 95.35  & 79.87  & 86.93  \\ 
        $\beta=0.05$ & 87.00  & 94.57  & 79.22  & 86.22  \\ 
        $\beta=0.1$ & \textbf{88.00}  & 95.38  & 80.52  & \textbf{87.32}  \\ 
        $\beta=0.2$ & 87.67  & 95.35  & 79.87  & 86.93  \\ 
        $\beta=0.25$ & 86.67  & 95.24  & 77.92  & 85.71  \\ 
        $\beta=0.5$ & 85.67  & 95.12  & 75.97  & 84.48 \\ 
    \bottomrule
    \end{tabular}
    }
    \vspace{-0.2cm}
    \caption{Ablation on hyper parameter $\beta$ selection.}
    \label{tab:abl-hyper2}
    \vspace{-0.4cm}
\end{table}

%% file: sup-case_study.tex
\begin{figure*}[tp]
    \centering
    \includegraphics[width=1\linewidth]{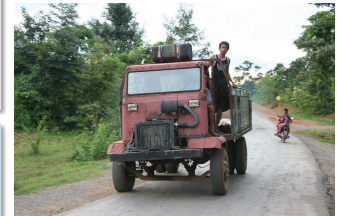}
        \caption{Illustration of hallucination correction by our proposed VaLiD with four more cases from AMBER \cite{AMBER} and POPE \cite{mscoco} dataset. Hallucinated objects from vanilla decoding are highlighted in \colorbox{yellow}{\textcolor{red}{\emph{\textbf{red}}}}.}
    \label{fig: sup-case_study}
    \vspace{-0.25cm}
\end{figure*}

%% file: pope-llava.tex
\begin{table*}[!ht]
    \centering
    \renewcommand\arraystretch{1.01}
    \setlength{\tabcolsep}{12pt}
    \resizebox{0.9\linewidth}{!}{%
    \begin{tabular}{cccccccc}
        \toprule
        \textbf{Dataset} & \textbf{Setup} &\textbf{Method} & Acc.$\uparrow$ & Precision & Recall & F1$\uparrow$ & Yes \\ \hline
        \multirow{15}{*}{\rotatebox{90}{\textbf{\normalsize MSCOCO \cite{mscoco}}}} & \multirow{5}{*}{\textit{Random}} & Vanilla & 82.33  & 89.50  & 73.27  & 80.57  & 40.93  \\ 
        ~ & ~ & VCD \cite{decodingVCD} & 87.23  & 93.87  & 79.67  & 86.19  & 42.43  \\ 
        ~ & ~ & M3ID \cite{decodingm3id} & 86.53  & 94.70  & 77.40  & 85.18  & 40.87  \\ 
        ~ & ~ & Ritual \cite{decodingritual} & 85.80  & 97.44  & 73.53  & 83.81  & 37.73  \\ 
        ~ & ~ & VaLiD & \textbf{89.03}  & 94.12  & 83.27  & \textbf{88.36}  & 42.47  \\ \cline{2-8}
        ~ & \multirow{5}{*}{\textit{Popular}} & Vanilla & 81.10  & 86.94  & 73.20  & 79.48  & 42.10  \\ 
        ~ & ~ & VCD \cite{decodingVCD} & 85.80  & 90.32  & 80.20  & 84.96  & 44.40  \\ 
        ~ & ~ & M3ID \cite{decodingm3id} & 84.63  & 90.75  & 77.13  & 83.39  & 42.50  \\ 
        ~ & ~ & Ritual \cite{decodingritual} & 84.93  & 94.11  & 74.53  & 83.18  & 39.60  \\ 
        ~ & ~ & VaLiD & \textbf{87.17}  & 90.31  & 83.27  & \textbf{86.65}  & 46.10  \\ \cline{2-8}
        ~ & \multirow{5}{*}{\textit{Adversarial}} & Vanilla & 78.63  & 81.06  & 74.73  & 77.76  & 46.10  \\ 
        ~ & ~ & VCD \cite{decodingVCD} & 81.40  & 83.36  & 78.47  & 80.84  & 47.07  \\ 
        ~ & ~ & M3ID \cite{decodingm3id} & 81.70  & 84.99  & 77.00  & 80.80  & 45.30  \\ 
        ~ & ~ & Ritual \cite{decodingritual} & 82.60  & 89.00  & 74.40  & 81.05  & 41.80  \\ 
        ~ & ~ & VaLiD & \textbf{83.20}  & 82.85  & 83.73  & \textbf{83.29}  & 50.53  \\ \midrule\midrule
        \multirow{15}{*}{\rotatebox{90}{\textbf{\normalsize A-OKVQA \cite{aokvqa}}}} & \multirow{5}{*}{\textit{Random}} & Vanilla & 85.53  & 87.69  & 82.67  & 85.11  & 47.13  \\ 
        ~ & ~ & VCD \cite{decodingVCD} & 87.90  & 88.26  & 87.20  & 87.72  & 49.40  \\ 
        ~ & ~ & M3ID \cite{decodingm3id} & 87.27  & 89.70  & 84.20  & 86.86  & 46.93  \\ 
        ~ & ~ & Ritual \cite{decodingritual} & 89.27  & 93.69  & 84.20  & 88.69  & 44.93  \\ 
        ~ & ~ & VaLiD & \textbf{90.53}  & 89.28  & 92.13  & \textbf{90.86}  & 51.60  \\ \cline{2-8}
        ~ & \multirow{5}{*}{\textit{Popular}} & Vanilla & 81.43  & 81.92  & 80.67  & 81.29  & 49.23  \\ 
        ~ & ~ & VCD \cite{decodingVCD} & 82.47  & 80.44  & 85.80  & 83.03  & 53.33  \\ 
        ~ & ~ & M3ID \cite{decodingm3id} & 83.53  & 83.76  & 83.20  & 83.48  & 49.67  \\ 
        ~ & ~ & Ritual \cite{decodingritual} & \textbf{86.60}  & 87.81  & 85.00  & 86.38  & 48.40  \\ 
        ~ & ~ & VaLiD & 86.17  & 82.20  & 92.33  & \textbf{86.97}  & 56.17  \\ \cline{2-8}
        ~ & \multirow{5}{*}{\textit{Adversarial}} & Vanilla & 75.47  & 72.82  & 81.27  & 76.81  & 55.80  \\ 
        ~ & ~ & VCD \cite{decodingVCD} & 76.60  & 72.41  & 85.93  & 78.60  & 59.33  \\ 
        ~ & ~ & M3ID \cite{decodingm3id} & 76.97  & 73.50  & 84.33  & 78.55  & 57.37  \\ 
        ~ & ~ & Ritual \cite{decodingritual} & \textbf{79.07}  & 77.08  & 82.73  & 79.81  & 53.67  \\ 
        ~ & ~ & VaLiD & 78.47  & 72.52  & 91.67  & \textbf{80.98}  & 63.20  \\ \midrule\midrule
        \multirow{15}{*}{\rotatebox{90}{\textbf{\normalsize GQA \cite{gqa}}}} & \multirow{5}{*}{\textit{Random}} & Vanilla & 84.70  & 88.19  & 80.13  & 83.97  & 45.43  \\ 
        ~ & ~ & VCD \cite{decodingVCD} & 87.90  & 89.40  & 86.00  & 87.67  & 48.10  \\ 
        ~ & ~ & M3ID \cite{decodingm3id} & 86.20  & 88.90  & 82.73  & 85.70  & 46.53  \\ 
        ~ & ~ & Ritual \cite{decodingritual} & 87.93  & 93.24  & 81.80  & 87.14  & 43.87  \\ 
        ~ & ~ & VaLiD & \textbf{89.73}  & 89.31  & 90.27  & \textbf{89.79}  & 50.53  \\ \cline{2-8}
        ~ & \multirow{5}{*}{\textit{Popular}} & Vanilla & 77.83  & 76.85  & 79.67  & 78.23  & 51.83  \\ 
        ~ & ~ & VCD \cite{decodingVCD} & 80.07  & 77.10  & 85.53  & 81.10  & 55.47  \\ 
        ~ & ~ & M3ID \cite{decodingm3id} & 78.60  & 77.97  & 79.73  & 78.84  & 51.13  \\ 
        ~ & ~ & Ritual \cite{decodingritual} & 82.08  & 83.65  & 81.53  & 82.58  & 48.73  \\ 
        ~ & ~ & VaLiD & \textbf{82.57}  & 78.19  & 90.33 & \textbf{83.82}  & 57.77  \\ \cline{2-8}
        ~ & \multirow{5}{*}{\textit{Adversarial}} & Vanilla & 75.70  & 73.93  & 79.40  & 76.57  & 53.70  \\ 
        ~ & ~ & VCD \cite{decodingVCD} & 76.30  & 72.63  & 84.40  & 78.08  & 58.10  \\ 
        ~ & ~ & M3ID \cite{decodingm3id} & 77.20  & 75.19  & 81.20  & 78.08  & 54.00  \\ 
        ~ & ~ & Ritual \cite{decodingritual} & \textbf{80.10}  & 79.00  & 82.00  & 80.47  & 51.90  \\ 
        ~ & ~ & VaLiD & 78.50  & 73.22  & 89.87  & \textbf{80.69}  & 61.37 \\ \bottomrule
    \end{tabular}
    }
    \caption{\textbf{Results of LLaVA-v1.5 on POPE benchmark}.}
    \label{tab:appendix-pope_llava}
    \vspace{-0.2cm}
\end{table*}

%% file: pope-blip2.tex
\begin{table*}[!ht]
    \centering
    \renewcommand\arraystretch{1.01}
    \setlength{\tabcolsep}{12pt}
    \resizebox{0.9\linewidth}{!}{%
    \begin{tabular}{cccccccc}
        \toprule
        \textbf{Dataset} & \textbf{Setup} &\textbf{Method} & Acc.$\uparrow$ & Precision & Recall & F1$\uparrow$ & Yes \\ \hline
        \multirow{15}{*}{\rotatebox{90}{\textbf{\normalsize MSCOCO \cite{mscoco}}}} & \multirow{5}{*}{\textit{Random}} & Vanilla & 82.20 & 83.45 & 80.33 & 81.86 & 48.13 \\ 
        ~ & ~ & VCD \cite{decodingVCD} & 84.37 & 86.95 & 80.87 & 83.80 & 46.50 \\ 
        ~ & ~ & M3ID \cite{decodingm3id} & 79.77 & 83.20 & 74.60 & 78.66 & 44.83 \\ 
        ~ & ~ & Ritual \cite{decodingritual} & 85.33 & 93.80 & 75.67 & \textbf{83.76} & 40.33 \\ 
        ~ & ~ & VaLiD & \textbf{85.90}  & 96.72  & 72.67  & 82.98  & 37.57 \\ \cline{2-8}
        ~ & \multirow{5}{*}{\textit{Popular}} & Vanilla & 79.10 & 78.55 & 80.07 & 79.30 & 50.97 \\ 
        ~ & ~ & VCD \cite{decodingVCD} & 81.23 & 82.24 & 79.67 & 80.93 & 48.43 \\ 
        ~ & ~ & M3ID \cite{decodingm3id} & 76.80 & 77.80 & 75.00 & 76.37 & 48.20 \\ 
        ~ & ~ & Ritual \cite{decodingritual} & 82.87 & 88.40 & 75.67 & 81.53 & 42.80 \\ 
        ~ & ~ & VaLiD & \textbf{83.43} & 91.82 & 73.40 & \textbf{81.59} & 39.97 \\ \cline{2-8}
        ~ & \multirow{5}{*}{\textit{Adversarial}} & Vanilla & 76.87 & 75.25 & 80.07 & 77.58 & 53.20 \\ 
        ~ & ~ & VCD \cite{decodingVCD} & 78.53 & 78.12 & 79.27 & 78.69 & 50.73 \\ 
        ~ & ~ & M3ID \cite{decodingm3id} & 74.93 & 73.79 & 77.33 & 75.52 & 52.40 \\ 
        ~ & ~ & Ritual \cite{decodingritual} & 80.77 & 84.26 & 75.67 & \textbf{79.73} & 44.90 \\ 
        ~ & ~ & VaLiD & \textbf{81.33} & 87.78 & 72.80 & 79.59 & 41.47 \\ \midrule\midrule
        \multirow{15}{*}{\rotatebox{90}{\textbf{\normalsize A-OKVQA \cite{aokvqa}}}} & \multirow{5}{*}{\textit{Random}} & Vanilla & 80.50 & 77.54 & 85.87 & 81.49 & 55.37 \\ 
        ~ & ~ & VCD \cite{decodingVCD} & 83.13 & 80.79 & 86.93 & 83.75 & 53.80 \\ 
        ~ & ~ & M3ID \cite{decodingm3id} & 78.53 & 79.00 & 77.73 & 78.36 & 49.20 \\ 
        ~ & ~ & Ritual \cite{decodingritual} & 84.53 & 88.31 & 79.60 & 83.73 & 45.07 \\ 
        ~ & ~ & VaLiD & \textbf{87.73} & 90.90 & 83.87 & \textbf{87.24} & 48.53 \\ \cline{2-8}
        ~ & \multirow{5}{*}{\textit{Popular}} & Vanilla & 76.57 & 72.55 & 85.47 & 78.48 & 58.90 \\ 
        ~ & ~ & VCD \cite{decodingVCD} & 78.76 & 74.76 & 86.87 & 80.36 & 58.10 \\ 
        ~ & ~ & M3ID \cite{decodingm3id} & 76.67 & 75.74 & 78.47 & 77.08 & 51.80 \\ 
        ~ & ~ & Ritual \cite{decodingritual} & 81.83 & 83.14 & 79.87 & 81.47 & 48.03 \\ 
        ~ & ~ & VaLiD & \textbf{83.07} & 82.80 & 83.47 & \textbf{83.13} & 50.40 \\ \cline{2-8}
        ~ & \multirow{5}{*}{\textit{Adversarial}} & Vanilla & 70.43 & 65.66 & 85.67 & 74.34 & 65.23 \\ 
        ~ & ~ & VCD \cite{decodingVCD} & 72.83 & 67.96 & 86.40 & 76.08 & 63.57 \\ 
        ~ & ~ & M3ID \cite{decodingm3id} & 68.23 & 65.15 & 78.40 & 71.16 & 60.17 \\ 
        ~ & ~ & Ritual \cite{decodingritual} & 74.63 & 72.49 & 79.40 & 75.79 & 54.77 \\ 
        ~ & ~ & VaLiD & \textbf{76.83} & 73.86 & 81.60 & \textbf{78.19} & 59.80 \\ \midrule\midrule
        \multirow{15}{*}{\rotatebox{90}{\textbf{\normalsize GQA \cite{gqa}}}} & \multirow{5}{*}{\textit{Random}} & Vanilla & 79.17 & 75.72 & 85.86 & 80.47 & 56.70 \\ 
        ~ & ~ & VCD \cite{decodingVCD} & 82.10 & 79.85 & 85.87 & 82.75 & 53.77 \\ 
        ~ & ~ & M3ID \cite{decodingm3id} & 78.37 & 79.08 & 77.13 & 78.10 & 48.77 \\ 
        ~ & ~ & Ritual \cite{decodingritual} & 83.26 & 87.69 & 77.40 & 82.23 & 44.13 \\ 
        ~ & ~ & VaLiD & \textbf{86.53} & 90.53 & 81.60 & \textbf{85.83} & 45.07 \\ \cline{2-8}
        ~ & \multirow{5}{*}{\textit{Popular}} & Vanilla & 74.23 & 69.98 & 84.87 & 76.71 & 60.63 \\ 
        ~ & ~ & VCD \cite{decodingVCD} & 76.63 & 72.51 & 85.80 & 78.60 & 59.17 \\ 
        ~ & ~ & M3ID \cite{decodingm3id} & 74.07 & 72.53 & 77.47 & 74.92 & 53.40 \\ 
        ~ & ~ & Ritual \cite{decodingritual} & 79.33 & 79.45 & 79.13 & 79.29 & 49.80 \\ 
        ~ & ~ & VaLiD & \textbf{80.10} & 79.26 & 81.53 & \textbf{80.38} & 51.43 \\ \cline{2-8}
        ~ & \multirow{5}{*}{\textit{Adversarial}} & Vanilla & 71.13 & 66.49 & 85.20 & 74.69 & 64.07 \\ 
        ~ & ~ & VCD \cite{decodingVCD} & 73.27 & 68.10 & 87.53 & 76.60 & 64.27 \\ 
        ~ & ~ & M3ID \cite{decodingm3id} & 68.80 & 66.26 & 76.60 & 71.06 & 57.80 \\ 
        ~ & ~ & Ritual \cite{decodingritual} & 74.67 & 72.64 & 79.13 & 75.75 & 54.47 \\ 
        ~ & ~ & VaLiD & \textbf{77.13} & 75.03 & 81.33 & 78.05 & 54.20 \\ \bottomrule
    \end{tabular}
    }
    \caption{\textbf{Results of InstructBLIP on POPE benchmark}.}
    \label{tab:appendix-pope_blip2}
    \vspace{-0.2cm}
\end{table*}

%% file: pope-qwen.tex
\begin{table*}[!ht]
    \centering
    \renewcommand\arraystretch{1.01}
    \setlength{\tabcolsep}{12pt}
    \resizebox{0.9\linewidth}{!}{%
    \begin{tabular}{cccccccc}
        \toprule
        \textbf{Dataset} & \textbf{Setup} &\textbf{Method} & Acc.$\uparrow$ & Precision & Recall & F1$\uparrow$ & Yes \\ \hline
        \multirow{15}{*}{\rotatebox{90}{\textbf{\normalsize MSCOCO \cite{mscoco}}}} & \multirow{5}{*}{\textit{Random}} & Vanilla & 85.17  & 96.56  & 72.93  & 83.10  & 37.77  \\ 
        ~ & ~ & VCD \cite{decodingVCD} & 87.17  & 97.13  & 76.60  & 85.65  & 39.43  \\ 
        ~ & ~ & M3ID \cite{decodingm3id} & 85.97  & 97.79  & 73.60  & 83.99  & 37.63  \\ 
        ~ & ~ & Ritual \cite{decodingritual} & 85.40  & 98.90  & 71.60  & 83.06  & 36.20  \\ 
        ~ & ~ & VaLiD & \textbf{89.07}  & 96.28  & 81.27  & \textbf{88.14}  & 42.20  \\ \cline{2-8}
        ~ & \multirow{5}{*}{\textit{Popular}} & Vanilla & 84.73  & 95.38  & 73.00  & 82.70  & 38.27  \\ 
        ~ & ~ & VCD \cite{decodingVCD} & 86.07  & 94.64  & 76.47  & 84.59  & 40.40  \\ 
        ~ & ~ & M3ID \cite{decodingm3id} & 84.87  & 95.40  & 73.27  & 82.88  & 38.40  \\ 
        ~ & ~ & Ritual \cite{decodingritual} & 84.84  & 97.20  & 71.73  & 82.55  & 36.90  \\ 
        ~ & ~ & VaLiD & \textbf{87.40} & 93.09  & 80.80  & \textbf{86.51}  & 43.40  \\ \cline{2-8}
        ~ & \multirow{5}{*}{\textit{Adversarial}} & Vanilla & 83.33  & 90.92  & 74.07  & 81.63  & 40.73  \\ 
        ~ & ~ & VCD \cite{decodingVCD} & 84.27  & 90.28  & 76.80  & 83.00  & 42.53  \\ 
        ~ & ~ & M3ID \cite{decodingm3id} & 83.23  & 91.16  & 73.60  & 81.45  & 40.37  \\ 
        ~ & ~ & Ritual \cite{decodingritual} & 82.80  & 93.16  & 70.80  & 80.45  & 38.00  \\ 
        ~ & ~ & VaLiD & \textbf{84.27}  & 86.35  & 81.40  & \textbf{83.80}  & 47.13  \\ \midrule\midrule
        \multirow{15}{*}{\rotatebox{90}{\textbf{\normalsize A-OKVQA \cite{aokvqa}}}} & \multirow{5}{*}{\textit{Random}} & Vanilla & 86.80  & 94.30  & 78.33  & 85.58  & 41.53  \\ 
        ~ & ~ & VCD \cite{decodingVCD} & 87.80  & 94.23  & 80.53  & 86.84  & 42.73  \\ 
        ~ & ~ & M3ID \cite{decodingm3id} & 87.43  & 95.54  & 78.53  & 86.21  & 41.10  \\ 
        ~ & ~ & Ritual \cite{decodingritual} & 86.10  & 96.09  & 75.27  & 84.41  & 39.16  \\ 
        ~ & ~ & VaLiD & \textbf{89.36}  & 91.89  & 86.40  & \textbf{89.04}  & 47.03  \\ \cline{2-8}
        ~ & \multirow{5}{*}{\textit{Popular}} & Vanilla & 86.73  & 95.16  & 77.40  & 85.37  & 40.67  \\ 
        ~ & ~ & VCD \cite{decodingVCD} & 87.53  & 93.17  & 81.00  & 86.66  & 43.47  \\ 
        ~ & ~ & M3ID \cite{decodingm3id} & 87.27  & 94.86  & 78.80  & 86.09  & 41.53  \\ 
        ~ & ~ & Ritual \cite{decodingritual} & 85.77  & 95.97  & 74.67  & 83.99  & 38.90  \\ 
        ~ & ~ & VaLiD & \textbf{88.53}  & 90.76  & 85.80  & \textbf{88.21}  & 47.27  \\ \cline{2-8}
        ~ & \multirow{5}{*}{\textit{Adversarial}} & Vanilla & 80.87  & 83.21  & 77.33  & 80.17  & 46.47  \\ 
        ~ & ~ & VCD \cite{decodingVCD} & 81.80  & 82.54  & 80.67  & 81.59  & 48.87  \\ 
        ~ & ~ & M3ID \cite{decodingm3id} & 81.40  & 83.50  & 78.27  & 80.80  & 46.87  \\ 
        ~ & ~ & Ritual \cite{decodingritual} & 80.83  & 84.85  & 75.07  & 79.66  & 44.23  \\ 
        ~ & ~ & VaLiD & \textbf{82.33}  & 79.86  & 86.47  & \textbf{83.03}  & 54.13  \\ \midrule\midrule
        \multirow{15}{*}{\rotatebox{90}{\textbf{\normalsize GQA \cite{gqa}}}} & \multirow{5}{*}{\textit{Random}} & Vanilla & 84.40  & 89.75  & 77.67  & 83.27  & 43.27  \\ 
        ~ & ~ & VCD \cite{decodingVCD} & 86.73  & 91.43  & 81.07  & 85.94  & 44.33  \\ 
        ~ & ~ & M3ID \cite{decodingm3id} & 85.93  & 91.40  & 79.33  & 84.94  & 43.40  \\ 
        ~ & ~ & Ritual \cite{decodingritual} & 85.80  & 93.52  & 76.93  & 84.42  & 41.13  \\ 
        ~ & ~ & VaLiD & \textbf{90.07}  & 92.15  & 87.60  & 89.81  & 47.53  \\ \cline{2-8}
        ~ & \multirow{5}{*}{\textit{Popular}} & Vanilla & 80.50  & 82.02  & 78.13  & 80.03  & 47.63  \\ 
        ~ & ~ & VCD \cite{decodingVCD} & 82.33  & 83.22  & 81.00  & 82.09  & 48.67  \\ 
        ~ & ~ & M3ID \cite{decodingm3id} & 81.07  & 83.72  & 77.13  & 80.29  & 46.07  \\ 
        ~ & ~ & Ritual \cite{decodingritual} & 82.56  & 87.84  & 75.60  & 81.26  & 43.03  \\ 
        ~ & ~ & VaLiD & \textbf{84.23}  & 82.44  & 87.00  & \textbf{84.66}  & 52.77  \\ \cline{2-8}
        ~ & \multirow{5}{*}{\textit{Adversarial}} & Vanilla & 79.33  & 79.81  & 78.53  & 79.17  & 49.20  \\ 
        ~ & ~ & VCD \cite{decodingVCD} & 81.10  & 81.08  & 81.13  & 81.11  & 50.03  \\ 
        ~ & ~ & M3ID \cite{decodingm3id} & 79.83  & 80.50  & 78.73  & 79.61  & 48.90  \\ 
        ~ & ~ & Ritual \cite{decodingritual} & 80.33  & 83.65  & 75.40  & 79.31  & 45.06  \\ 
        ~ & ~ & VaLiD & \textbf{82.40}  & 79.42  & 87.47  & \textbf{83.25}  & 55.07 \\ \bottomrule
    \end{tabular}
    }
    \caption{\textbf{Results of Qwen-VL on POPE benchmark}.}
    \label{tab:appendix-pope_qwen}
    \vspace{-0.2cm}
\end{table*}

%% file: main.bbl
\begin{thebibliography}{61}
\providecommand{\natexlab}[1]{#1}
\providecommand{\url}[1]{\texttt{#1}}
\expandafter\ifx\csname urlstyle\endcsname\relax
  \providecommand{\doi}[1]{doi: #1}\else
  \providecommand{\doi}{doi: \begingroup \urlstyle{rm}\Url}\fi

\bibitem[Antol et~al.(2015)Antol, Agrawal, Lu, Mitchell, Batra, Zitnick, and Parikh]{vqa}
Stanislaw Antol, Aishwarya Agrawal, Jiasen Lu, Margaret Mitchell, Dhruv Batra, C~Lawrence Zitnick, and Devi Parikh.
\newblock Vqa: Visual question answering.
\newblock In \emph{Proceedings of the IEEE international conference on computer vision}, pages 2425--2433, 2015.

\bibitem[Bai et~al.(2023)Bai, Bai, Yang, Wang, Tan, Wang, Lin, Zhou, and Zhou]{qwenvl}
Jinze Bai, Shuai Bai, Shusheng Yang, Shijie Wang, Sinan Tan, Peng Wang, Junyang Lin, Chang Zhou, and Jingren Zhou.
\newblock Qwen-vl: A versatile vision-language model for understanding, localization, text reading, and beyond, 2023.

\bibitem[Bubeck et~al.(2023)Bubeck, Chandrasekaran, Eldan, Gehrke, Horvitz, Kamar, Lee, Lee, Li, Lundberg, et~al.]{AGI}
S{\'e}bastien Bubeck, Varun Chandrasekaran, Ronen Eldan, Johannes Gehrke, Eric Horvitz, Ece Kamar, Peter Lee, Yin~Tat Lee, Yuanzhi Li, Scott Lundberg, et~al.
\newblock Sparks of artificial general intelligence: Early experiments with gpt-4.
\newblock \emph{arXiv preprint arXiv:2303.12712}, 2023.

\bibitem[Chen et~al.(2024{\natexlab{a}})Chen, Xiong, Liu, Wu, Xiao, Gao, and He]{uncertainty2}
Shiqi Chen, Miao Xiong, Junteng Liu, Zhengxuan Wu, Teng Xiao, Siyang Gao, and Junxian He.
\newblock In-context sharpness as alerts: An inner representation perspective for hallucination mitigation.
\newblock \emph{arXiv preprint arXiv:2403.01548}, 2024{\natexlab{a}}.

\bibitem[Chen et~al.(2024{\natexlab{b}})Chen, Zhao, Luo, Yao, Li, and Zhou]{2.1_11HALC}
Zhaorun Chen, Zhuokai Zhao, Hongyin Luo, Huaxiu Yao, Bo Li, and Jiawei Zhou.
\newblock Halc: Object hallucination reduction via adaptive focal-contrast decoding.
\newblock \emph{arXiv preprint arXiv:2403.00425}, 2024{\natexlab{b}}.

\bibitem[Chuang et~al.(2024)Chuang, Xie, Luo, Kim, Glass, and He]{DOLA}
Yung-Sung Chuang, Yujia Xie, Hongyin Luo, Yoon Kim, James Glass, and Pengcheng He.
\newblock Dola: Decoding by contrasting layers improves factuality in large language models, 2024.

\bibitem[Dai et~al.(2023)Dai, Li, Li, Tiong, Zhao, Wang, Li, Fung, and Hoi]{instructblip}
Wenliang Dai, Junnan Li, Dongxu Li, Anthony Meng~Huat Tiong, Junqi Zhao, Weisheng Wang, Boyang Li, Pascale Fung, and Steven Hoi.
\newblock Instructblip: Towards general-purpose vision-language models with instruction tuning, 2023.

\bibitem[Deng et~al.(2024)Deng, Chen, and Hooi]{decodingclip}
Ailin Deng, Zhirui Chen, and Bryan Hooi.
\newblock Seeing is believing: Mitigating hallucination in large vision-language models via clip-guided decoding, 2024.

\bibitem[Fang et~al.(2022)Fang, Wang, Xie, Sun, Wu, Wang, Huang, Wang, and Cao]{evavit}
Yuxin Fang, Wen Wang, Binhui Xie, Quan Sun, Ledell Wu, Xinggang Wang, Tiejun Huang, Xinlong Wang, and Yue Cao.
\newblock Eva: Exploring the limits of masked visual representation learning at scale, 2022.

\bibitem[Favero et~al.(2024{\natexlab{a}})Favero, Zancato, Trager, Choudhary, Perera, Achille, Swaminathan, and Soatto]{decodingm3id}
Alessandro Favero, Luca Zancato, Matthew Trager, Siddharth Choudhary, Pramuditha Perera, Alessandro Achille, Ashwin Swaminathan, and Stefano Soatto.
\newblock Multi-modal hallucination control by visual information grounding.
\newblock In \emph{Proceedings of the IEEE/CVF Conference on Computer Vision and Pattern Recognition}, pages 14303--14312, 2024{\natexlab{a}}.

\bibitem[Favero et~al.(2024{\natexlab{b}})Favero, Zancato, Trager, Choudhary, Perera, Achille, Swaminathan, and Soatto]{languagepri5}
Alessandro Favero, Luca Zancato, Matthew Trager, Siddharth Choudhary, Pramuditha Perera, Alessandro Achille, Ashwin Swaminathan, and Stefano Soatto.
\newblock Multi-modal hallucination control by visual information grounding.
\newblock In \emph{Proceedings of the IEEE/CVF Conference on Computer Vision and Pattern Recognition}, pages 14303--14312, 2024{\natexlab{b}}.

\bibitem[Fu et~al.(2024)Fu, Chen, Shen, Qin, Zhang, Lin, Yang, Zheng, Li, Sun, Wu, and Ji]{mme}
Chaoyou Fu, Peixian Chen, Yunhang Shen, Yulei Qin, Mengdan Zhang, Xu Lin, Jinrui Yang, Xiawu Zheng, Ke Li, Xing Sun, Yunsheng Wu, and Rongrong Ji.
\newblock Mme: A comprehensive evaluation benchmark for multimodal large language models, 2024.

\bibitem[Gunjal et~al.(2024{\natexlab{a}})Gunjal, Yin, and Bas]{fine-tuning1}
Anisha Gunjal, Jihan Yin, and Erhan Bas.
\newblock Detecting and preventing hallucinations in large vision language models.
\newblock In \emph{Proceedings of the AAAI Conference on Artificial Intelligence}, pages 18135--18143, 2024{\natexlab{a}}.

\bibitem[Gunjal et~al.(2024{\natexlab{b}})Gunjal, Yin, and Bas]{introhal1}
Anisha Gunjal, Jihan Yin, and Erhan Bas.
\newblock Detecting and preventing hallucinations in large vision language models.
\newblock In \emph{Proceedings of the AAAI Conference on Artificial Intelligence}, pages 18135--18143, 2024{\natexlab{b}}.

\bibitem[He et~al.(2024)He, Wei, Xie, and Tian]{2.1_6}
Xin He, Longhui Wei, Lingxi Xie, and Qi Tian.
\newblock Incorporating visual experts to resolve the information loss in multimodal large language models.
\newblock \emph{arXiv preprint arXiv:2401.03105}, 2024.

\bibitem[Huang et~al.(2023)Huang, Yu, Ma, Zhong, Feng, Wang, Chen, Peng, Feng, Qin, and Liu]{introhal3}
Lei Huang, Weijiang Yu, Weitao Ma, Weihong Zhong, Zhangyin Feng, Haotian Wang, Qianglong Chen, Weihua Peng, Xiaocheng Feng, Bing Qin, and Ting Liu.
\newblock A survey on hallucination in large language models: Principles, taxonomy, challenges, and open questions, 2023.

\bibitem[Huang et~al.(2024)Huang, Dong, Zhang, Wang, He, Wang, Lin, Zhang, and Yu]{OPERA}
Qidong Huang, Xiaoyi Dong, Pan Zhang, Bin Wang, Conghui He, Jiaqi Wang, Dahua Lin, Weiming Zhang, and Nenghai Yu.
\newblock Opera: Alleviating hallucination in multi-modal large language models via over-trust penalty and retrospection-allocation.
\newblock In \emph{Proceedings of the IEEE/CVF Conference on Computer Vision and Pattern Recognition}, pages 13418--13427, 2024.

\bibitem[Hudson and Manning(2019)]{gqa}
Drew~A Hudson and Christopher~D Manning.
\newblock Gqa: A new dataset for real-world visual reasoning and compositional question answering.
\newblock In \emph{Proceedings of the IEEE/CVF conference on computer vision and pattern recognition}, pages 6700--6709, 2019.

\bibitem[Jiang et~al.(2024)Jiang, Xu, Dong, Chen, Ye, Yan, Ye, Zhang, Huang, and Zhang]{2.1_1}
Chaoya Jiang, Haiyang Xu, Mengfan Dong, Jiaxing Chen, Wei Ye, Ming Yan, Qinghao Ye, Ji Zhang, Fei Huang, and Shikun Zhang.
\newblock Hallucination augmented contrastive learning for multimodal large language model.
\newblock In \emph{Proceedings of the IEEE/CVF Conference on Computer Vision and Pattern Recognition}, pages 27036--27046, 2024.

\bibitem[Jiao et~al.(2024)Jiao, Chen, Huang, Li, and Shen]{2.1_7}
Qirui Jiao, Daoyuan Chen, Yilun Huang, Yaliang Li, and Ying Shen.
\newblock Enhancing multimodal large language models with vision detection models: An empirical study.
\newblock \emph{arXiv preprint arXiv:2401.17981}, 2024.

\bibitem[Kim et~al.(2024)Kim, Cho, Bae, Ahn, and Yun]{kim2024vacode}
Sihyeon Kim, Boryeong Cho, Sangmin Bae, Sumyeong Ahn, and Se-Young Yun.
\newblock Vacode: Visual augmented contrastive decoding.
\newblock \emph{arXiv preprint arXiv:2408.05337}, 2024.

\bibitem[Leng et~al.(2024)Leng, Zhang, Chen, Li, Lu, Miao, and Bing]{decodingVCD}
Sicong Leng, Hang Zhang, Guanzheng Chen, Xin Li, Shijian Lu, Chunyan Miao, and Lidong Bing.
\newblock Mitigating object hallucinations in large vision-language models through visual contrastive decoding.
\newblock In \emph{Proceedings of the IEEE/CVF Conference on Computer Vision and Pattern Recognition}, pages 13872--13882, 2024.

\bibitem[Li et~al.(2025)Li, Zhang, Jie, Ma, and Li]{IMCCD}
Jiaming Li, Jiacheng Zhang, Zequn Jie, Lin Ma, and Guanbin Li.
\newblock Mitigating hallucination for large vision language model by inter-modality correlation calibration decoding.
\newblock \emph{arXiv preprint arXiv:2501.01926}, 2025.

\bibitem[Li et~al.(2023{\natexlab{a}})Li, Yin, Li, Chen, Wang, Ren, Li, Yang, Xu, Sun, et~al.]{fine-tuning2}
Lei Li, Yuwei Yin, Shicheng Li, Liang Chen, Peiyi Wang, Shuhuai Ren, Mukai Li, Yazheng Yang, Jingjing Xu, Xu Sun, et~al.
\newblock A large-scale dataset towards multi-modal multilingual instruction tuning.
\newblock \emph{arXiv preprint arXiv:2306.04387}, 3, 2023{\natexlab{a}}.

\bibitem[Li et~al.(2022)Li, Holtzman, Fried, Liang, Eisner, Hashimoto, Zettlemoyer, and Lewis]{2.2_1}
Xiang~Lisa Li, Ari Holtzman, Daniel Fried, Percy Liang, Jason Eisner, Tatsunori Hashimoto, Luke Zettlemoyer, and Mike Lewis.
\newblock Contrastive decoding: Open-ended text generation as optimization.
\newblock \emph{arXiv preprint arXiv:2210.15097}, 2022.

\bibitem[Li et~al.(2023{\natexlab{b}})Li, Du, Zhou, Wang, Zhao, and Wen]{introhal2}
Yifan Li, Yifan Du, Kun Zhou, Jinpeng Wang, Wayne~Xin Zhao, and Ji-Rong Wen.
\newblock Evaluating object hallucination in large vision-language models.
\newblock \emph{arXiv preprint arXiv:2305.10355}, 2023{\natexlab{b}}.

\bibitem[Li et~al.(2023{\natexlab{c}})Li, Du, Zhou, Wang, Zhao, and Wen]{pope}
Yifan Li, Yifan Du, Kun Zhou, Jinpeng Wang, Wayne~Xin Zhao, and Ji-Rong Wen.
\newblock Evaluating object hallucination in large vision-language models.
\newblock \emph{arXiv preprint arXiv:2305.10355}, 2023{\natexlab{c}}.

\bibitem[Lin et~al.(2014)Lin, Maire, Belongie, Hays, Perona, Ramanan, Doll{\'a}r, and Zitnick]{mscoco}
Tsung-Yi Lin, Michael Maire, Serge Belongie, James Hays, Pietro Perona, Deva Ramanan, Piotr Doll{\'a}r, and C~Lawrence Zitnick.
\newblock Microsoft coco: Common objects in context.
\newblock In \emph{Computer Vision--ECCV 2014: 13th European Conference, Zurich, Switzerland, September 6-12, 2014, Proceedings, Part V 13}, pages 740--755. Springer, 2014.

\bibitem[Liu et~al.(2021)Liu, Sap, Lu, Swayamdipta, Bhagavatula, Smith, and Choi]{CD2}
Alisa Liu, Maarten Sap, Ximing Lu, Swabha Swayamdipta, Chandra Bhagavatula, Noah~A. Smith, and Yejin Choi.
\newblock Dexperts: Decoding-time controlled text generation with experts and anti-experts, 2021.

\bibitem[Liu et~al.(2023{\natexlab{a}})Liu, Lin, Li, Wang, Yacoob, and Wang]{LRV-Instruction}
Fuxiao Liu, Kevin Lin, Linjie Li, Jianfeng Wang, Yaser Yacoob, and Lijuan Wang.
\newblock Mitigating hallucination in large multi-modal models via robust instruction tuning.
\newblock In \emph{The Twelfth International Conference on Learning Representations}, 2023{\natexlab{a}}.

\bibitem[Liu et~al.(2023{\natexlab{b}})Liu, Lin, Li, Wang, Yacoob, and Wang]{fine-tuning3}
Fuxiao Liu, Kevin Lin, Linjie Li, Jianfeng Wang, Yaser Yacoob, and Lijuan Wang.
\newblock Aligning large multi-modal model with robust instruction tuning.
\newblock \emph{arXiv preprint arXiv:2306.14565}, 2023{\natexlab{b}}.

\bibitem[Liu et~al.(2023{\natexlab{c}})Liu, Li, Wu, and Lee]{visualinstructiontuning}
Haotian Liu, Chunyuan Li, Qingyang Wu, and Yong~Jae Lee.
\newblock Visual instruction tuning, 2023{\natexlab{c}}.

\bibitem[Liu et~al.(2024{\natexlab{a}})Liu, Li, Li, and Lee]{llava1.5}
Haotian Liu, Chunyuan Li, Yuheng Li, and Yong~Jae Lee.
\newblock Improved baselines with visual instruction tuning, 2024{\natexlab{a}}.

\bibitem[Liu et~al.(2024{\natexlab{b}})Liu, Xue, Chen, Chen, Zhao, Wang, Hou, Li, and Peng]{survey3}
Hanchao Liu, Wenyuan Xue, Yifei Chen, Dapeng Chen, Xiutian Zhao, Ke Wang, Liping Hou, Rongjun Li, and Wei Peng.
\newblock A survey on hallucination in large vision-language models.
\newblock \emph{arXiv preprint arXiv:2402.00253}, 2024{\natexlab{b}}.

\bibitem[Lu et~al.(2024)Lu, Rao, Chen, Guo, Zhang, Sun, Yang, and Yang]{2.1_3}
Jiaying Lu, Jinmeng Rao, Kezhen Chen, Xiaoyuan Guo, Yawen Zhang, Baochen Sun, Carl Yang, and Jie Yang.
\newblock Evaluation and enhancement of semantic grounding in large vision-language models.
\newblock In \emph{AAAI-ReLM Workshop}, 2024.

\bibitem[O'Brien and Lewis(2023)]{2.2_2}
Sean O'Brien and Mike Lewis.
\newblock Contrastive decoding improves reasoning in large language models.
\newblock \emph{arXiv preprint arXiv:2309.09117}, 2023.

\bibitem[Qiu et~al.(2024)Qiu, Ou, Wu, Li, Liu, and King]{uncertainty6}
Zexuan Qiu, Zijing Ou, Bin Wu, Jingjing Li, Aiwei Liu, and Irwin King.
\newblock Entropy-based decoding for retrieval-augmented large language models.
\newblock \emph{arXiv preprint arXiv:2406.17519}, 2024.

\bibitem[Radford et~al.(2021)Radford, Kim, Hallacy, Ramesh, Goh, Agarwal, Sastry, Askell, Mishkin, Clark, Krueger, and Sutskever]{clip}
Alec Radford, Jong~Wook Kim, Chris Hallacy, Aditya Ramesh, Gabriel Goh, Sandhini Agarwal, Girish Sastry, Amanda Askell, Pamela Mishkin, Jack Clark, Gretchen Krueger, and Ilya Sutskever.
\newblock Learning transferable visual models from natural language supervision, 2021.

\bibitem[Rawte et~al.(2023)Rawte, Sheth, and Das]{introhal4}
Vipula Rawte, Amit Sheth, and Amitava Das.
\newblock A survey of hallucination in large foundation models.
\newblock \emph{arXiv preprint arXiv:2309.05922}, 2023.

\bibitem[Schwenk et~al.(2022)Schwenk, Khandelwal, Clark, Marino, and Mottaghi]{aokvqa}
Dustin Schwenk, Apoorv Khandelwal, Christopher Clark, Kenneth Marino, and Roozbeh Mottaghi.
\newblock A-okvqa: A benchmark for visual question answering using world knowledge.
\newblock In \emph{European conference on computer vision}, pages 146--162. Springer, 2022.

\bibitem[Sun et~al.(2023)Sun, Shen, Cao, Liu, Li, Shen, Gan, Gui, Wang, Yang, et~al.]{2.1_4}
Zhiqing Sun, Sheng Shen, Shengcao Cao, Haotian Liu, Chunyuan Li, Yikang Shen, Chuang Gan, Liang-Yan Gui, Yu-Xiong Wang, Yiming Yang, et~al.
\newblock Aligning large multimodal models with factually augmented rlhf.
\newblock \emph{arXiv preprint arXiv:2309.14525}, 2023.

\bibitem[Wan et~al.(2024)Wan, Cho, Stengel-Eskin, and Bansal]{post3}
David Wan, Jaemin Cho, Elias Stengel-Eskin, and Mohit Bansal.
\newblock Contrastive region guidance: Improving grounding in vision-language models without training.
\newblock \emph{arXiv preprint arXiv:2403.02325}, 2024.

\bibitem[Wang et~al.(2024{\natexlab{a}})Wang, Wu, Han, Peng, Zhong, Zhang, Dong, Li, Li, Wang, et~al.]{2.1_5}
Bin Wang, Fan Wu, Xiao Han, Jiahui Peng, Huaping Zhong, Pan Zhang, Xiaoyi Dong, Weijia Li, Wei Li, Jiaqi Wang, et~al.
\newblock Vigc: Visual instruction generation and correction.
\newblock In \emph{Proceedings of the AAAI Conference on Artificial Intelligence}, pages 5309--5317, 2024{\natexlab{a}}.

\bibitem[Wang et~al.(2023)Wang, Wang, Xu, Zhang, Gu, Jia, Yan, Zhang, and Sang]{AMBER}
Junyang Wang, Yuhang Wang, Guohai Xu, Jing Zhang, Yukai Gu, Haitao Jia, Ming Yan, Ji Zhang, and Jitao Sang.
\newblock An llm-free multi-dimensional benchmark for mllms hallucination evaluation.
\newblock \emph{arXiv preprint arXiv:2311.07397}, 2023.

\bibitem[Wang et~al.()Wang, Gu, Gao, and Zhou]{damo}
Kaishen Wang, Hengrui Gu, Meijun Gao, and Kaixiong Zhou.
\newblock Damo: Decoding by accumulating activations momentum for mitigating hallucinations in vision-language models.
\newblock In \emph{The Thirteenth International Conference on Learning Representations}.

\bibitem[Wang et~al.(2016)Wang, Yin, Wang, Wu, and Wang]{CrossmodalRetrieval}
Kaiye Wang, Qiyue Yin, Wei Wang, Shu Wu, and Liang Wang.
\newblock A comprehensive survey on cross-modal retrieval.
\newblock \emph{arXiv preprint arXiv:1607.06215}, 2016.

\bibitem[Wang et~al.(2024{\natexlab{b}})Wang, Pan, Ding, and Biemann]{decodingICD}
Xintong Wang, Jingheng Pan, Liang Ding, and Chris Biemann.
\newblock Mitigating hallucinations in large vision-language models with instruction contrastive decoding.
\newblock \emph{arXiv preprint arXiv:2403.18715}, 2024{\natexlab{b}}.

\bibitem[Woo et~al.(2024)Woo, Jang, Kim, Choi, and Kim]{decodingritual}
Sangmin Woo, Jaehyuk Jang, Donguk Kim, Yubin Choi, and Changick Kim.
\newblock Ritual: Random image transformations as a universal anti-hallucination lever in lvlms.
\newblock \emph{arXiv preprint arXiv:2405.17821}, 2024.

\bibitem[Wu et~al.(2024)Wu, Liu, Wang, Zhang, Wu, Wang, and Tan]{2.1_10}
Junfei Wu, Qiang Liu, Ding Wang, Jinghao Zhang, Shu Wu, Liang Wang, and Tieniu Tan.
\newblock Logical closed loop: Uncovering object hallucinations in large vision-language models.
\newblock \emph{arXiv preprint arXiv:2402.11622}, 2024.

\bibitem[Wu and Xie(2024)]{2.1_9}
Penghao Wu and Saining Xie.
\newblock V?: Guided visual search as a core mechanism in multimodal llms.
\newblock In \emph{Proceedings of the IEEE/CVF Conference on Computer Vision and Pattern Recognition}, pages 13084--13094, 2024.

\bibitem[Xiao and Wang(2021{\natexlab{a}})]{uncertainty1}
Yijun Xiao and William~Yang Wang.
\newblock On hallucination and predictive uncertainty in conditional language generation.
\newblock \emph{arXiv preprint arXiv:2103.15025}, 2021{\natexlab{a}}.

\bibitem[Xiao and Wang(2021{\natexlab{b}})]{uncertainty5}
Yijun Xiao and William~Yang Wang.
\newblock On hallucination and predictive uncertainty in conditional language generation.
\newblock \emph{arXiv preprint arXiv:2103.15025}, 2021{\natexlab{b}}.

\bibitem[Yin et~al.(2023)Yin, Fu, Zhao, Xu, Wang, Sui, Shen, Li, Sun, and Chen]{post2}
Shukang Yin, Chaoyou Fu, Sirui Zhao, Tong Xu, Hao Wang, Dianbo Sui, Yunhang Shen, Ke Li, Xing Sun, and Enhong Chen.
\newblock Woodpecker: Hallucination correction for multimodal large language models.
\newblock \emph{arXiv preprint arXiv:2310.16045}, 2023.

\bibitem[Zhai et~al.(2023)Zhai, Yang, Xu, Shen, Keutzer, and Li]{2.1_8}
Bohan Zhai, Shijia Yang, Chenfeng Xu, Sheng Shen, Kurt Keutzer, and Manling Li.
\newblock Halle-switch: Controlling object hallucination in large vision language models.
\newblock \emph{arXiv e-prints}, pages arXiv--2310, 2023.

\bibitem[Zhang et~al.(2023)Zhang, Qiu, Guo, Deng, Zhang, Zhang, Zhou, Wang, and Fu]{uncertainty4}
Tianhang Zhang, Lin Qiu, Qipeng Guo, Cheng Deng, Yue Zhang, Zheng Zhang, Chenghu Zhou, Xinbing Wang, and Luoyi Fu.
\newblock Enhancing uncertainty-based hallucination detection with stronger focus.
\newblock \emph{arXiv preprint arXiv:2311.13230}, 2023.

\bibitem[Zhang et~al.(2024)Zhang, Yu, Wen, Wang, Zhang, Wang, Jin, and Tan]{languagepri6}
Yi-Fan Zhang, Weichen Yu, Qingsong Wen, Xue Wang, Zhang Zhang, Liang Wang, Rong Jin, and Tieniu Tan.
\newblock Debiasing large visual language models.
\newblock \emph{arXiv preprint arXiv:2403.05262}, 2024.

\bibitem[Zhao et~al.(2024{\natexlab{a}})Zhao, Deng, Zhang, and Gu]{MARINE}
Linxi Zhao, Yihe Deng, Weitong Zhang, and Quanquan Gu.
\newblock Mitigating object hallucination in large vision-language models via classifier-free guidance.
\newblock \emph{arXiv preprint arXiv:2402.08680}, 2024{\natexlab{a}}.

\bibitem[Zhao et~al.(2024{\natexlab{b}})Zhao, Deng, Zhang, and Gu]{post4}
Linxi Zhao, Yihe Deng, Weitong Zhang, and Quanquan Gu.
\newblock Mitigating object hallucination in large vision-language models via classifier-free guidance.
\newblock \emph{arXiv preprint arXiv:2402.08680}, 2024{\natexlab{b}}.

\bibitem[Zhao et~al.(2024{\natexlab{c}})Zhao, Monti, Lehmann, and Assem]{2.2_3}
Zheng Zhao, Emilio Monti, Jens Lehmann, and Haytham Assem.
\newblock Enhancing contextual understanding in large language models through contrastive decoding.
\newblock In \emph{NAACL 2024}, 2024{\natexlab{c}}.

\bibitem[Zhou et~al.(2023)Zhou, Cui, Yoon, Zhang, Deng, Finn, Bansal, and Yao]{post1}
Yiyang Zhou, Chenhang Cui, Jaehong Yoon, Linjun Zhang, Zhun Deng, Chelsea Finn, Mohit Bansal, and Huaxiu Yao.
\newblock Analyzing and mitigating object hallucination in large vision-language models.
\newblock \emph{arXiv preprint arXiv:2310.00754}, 2023.

\bibitem[Zhu et~al.(2024)Zhu, Ji, Chen, Xu, Ye, and Liu]{IBD}
Lanyun Zhu, Deyi Ji, Tianrun Chen, Peng Xu, Jieping Ye, and Jun Liu.
\newblock Ibd: Alleviating hallucinations in large vision-language models via image-biased decoding, 2024.

\end{thebibliography}
